\documentclass[journal]{IEEEtran}

\ifCLASSINFOpdf
  \usepackage[pdftex]{graphicx}
  \graphicspath{{../pdf/}{../jpeg/}}
  \DeclareGraphicsExtensions{.pdf,.jpeg,.png}
\else
\fi

\usepackage{amsmath}

\usepackage{amssymb}  %
\usepackage{leftidx}
\usepackage{multirow}

\usepackage{subcaption}

\usepackage{algpseudocode}
\usepackage{algorithm}

\makeatletter
\let\NAT@parse\undefined
\makeatother
\usepackage[pagebackref=true,bookmarks=false,hidelinks]{hyperref}

\usepackage{xspace}

\makeatletter
\DeclareRobustCommand\onedot{\futurelet\@let@token\@onedot}
\def\@onedot{\ifx\@let@token.\else.\null\fi\xspace}

\def\eg{\emph{e.g}\onedot} 
\def\ie{\emph{i.e}\onedot} 
 
 \def\vs{\emph{vs}\onedot}
 
\def\etal{\emph{et al}\onedot}
\makeatother
\DeclareMathAlphabet{\mbf}{OT1}{ptm}{b}{n}
\newcommand{\mbc}[1]{{{\boldsymbol{\mathcal{#1}}}}}

\newcommand*\numcircledmod[1]{\raisebox{.5pt}{\textcircled{\raisebox{-.9pt} {#1}}}}

\usepackage{xcolor}

\begin{document}
\title{A Versatile Keyframe-Based Structureless Filter \\
	 for Visual Inertial Odometry}

\author{Jianzhu~Huai$^{1}$,
        Yukai~Lin$^{2}$,
        Charles~Toth$^{3}$,~\IEEEmembership{Senior Member,~IEEE},\\
        Yuan~Zhuang$^{1\dagger}$,~\IEEEmembership{Member,~IEEE},
        and~Dong~Chen$^{1}$,~\IEEEmembership{Member,~IEEE}
\thanks{$^{1}$J. Huai, Y. Zhuang, D. Chen are with the State Key
Laboratory of Surveying, Mapping and Remote Sensing, Wuhan University,
129 Luoyu Road, Wuhan, Hubei, 430079, China. 
Homepage of J. Huai: https://www.jianzhuhuai.com/.}%
\thanks{$^{2}$Y. Lin is with ETH Zurich, Switzerland.}%
\thanks{$^{3}$Charles Toth is with the Department
	of Civil, Environmental, and Geodetic Engineering, The Ohio State University, Columbus,
	OH, USA.}%
\thanks{$^{\dagger}$Corresponding author, yuan.zhuang@whu.edu.cn.}%
}

\maketitle

\begin{abstract}
Motion estimation by fusing data from at least a camera and an Inertial
Measurement Unit (IMU) enables many applications in robotics.
However, among the multitude of Visual Inertial Odometry (VIO) methods,
few efficiently estimate device motion with consistent
covariance, and calibrate sensor
parameters online for handling data from consumer sensors.
This paper addresses the gap with a Keyframe-based Structureless Filter
(KSF).
For efficiency, landmarks are not included in the filter's state vector.
For robustness, KSF associates feature observations and manages state
variables using the concept of keyframes.
For flexibility, KSF supports anytime calibration of IMU systematic errors,
as well as extrinsic, intrinsic, and temporal parameters of each camera.
Estimator consistency and observability of sensor parameters were analyzed by simulation.
Sensitivity to design options, \eg, feature matching method and camera count was studied with the EuRoC benchmark.
Sensor parameter estimation was evaluated on raw TUM VI
sequences and smartphone data.
Moreover, pose estimation accuracy was evaluated on EuRoC and TUM VI
sequences versus recent VIO methods.
These tests confirm that KSF reliably calibrates sensor parameters when the data
contain adequate motion, and consistently estimate
motion with accuracy rivaling recent VIO methods.
Our implementation runs at 42 Hz with stereo camera images on a consumer laptop.
\end{abstract}


\IEEEpeerreviewmaketitle

\section*{Supplementary Material}

A video of running the KSF method with real-time calibration and consistent
pose covariance on TUM VI raw room1 sequence is
available at \href{https://youtu.be/ZJFTdtrP2HQ}{https://youtu.be/ZJFTdtrP2HQ}.

\section{Introduction}
\IEEEPARstart{V}{isual} Inertial Odometry (VIO) methods estimate the motion of a platform outfitted
with at least a camera and an Inertial Measurement Unit (IMU).
They are an enabling technique for augmented reality and robotics.
For instance, these methods have been widely used for state estimation 
of smartphones \cite{arkit_2020}, Unmanned Aerial Vehicles (UAVs) \cite{delmerico2018benchmark},
and service robots \cite{huai_segway_2019}, thanks to the low weight and power 
consumption and the wide availability of the sensor assembly.
As the VIO technology matures, a plethora of VIO algorithms \cite{cadena2016past} have
been proposed over the years.
They come in a variety of frontend feature association schemes and backend estimation engines.
The variants of feature association schemes mainly revolve around whether feature descriptors are extracted (\eg, \cite{mur-artal_2015} \vs \cite{engel_direct_2018}) and
whether feature matching is done between frames and selected frames \ie, keyframes 
(\eg., \cite{mur-artal_2015} \vs \cite{geneva_openvins_2019}).
The backend estimation engines are mainly based on filtering, optimization, neural networks, or hybrids.
To reduce computation time of jointly solving for ego-motion and scene structure in an estimator,
a structureless approach can be taken where the estimator only solves for ego-motion and the scene structure is solved separately.

Despite the rich literature on VIO, there remain some desired properties of a VIO method that are not well addressed.
Ideally, a VIO method should produce a covariance estimate for the
estimated pose (position and orientation). But for some
optimization-based methods, it is computationally intensive
to obtain the covariance \cite{leutenegger_keyframe_2015}, 
while some filtering-based methods report over-confident covariance
inconsistent to the actual estimation error \cite{li_high_2013}.
Lack of consistent covariance further prevents proper weighting of
pose estimates in downstream algorithms, \eg, pose graph optimization \cite{rosinolKimera2020}.

\begin{figure}[!t]
	\centering
	\includegraphics[width=\columnwidth]{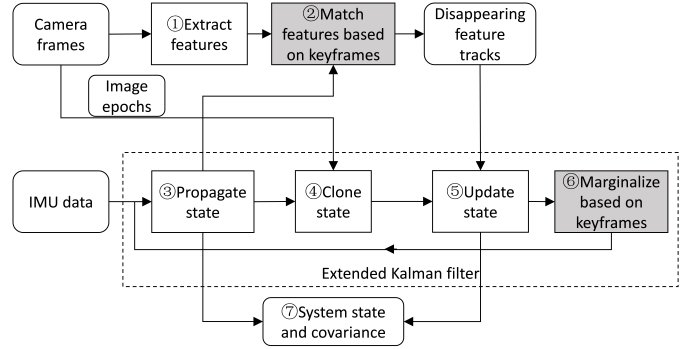}
	\caption{Flowchart of the keyframe-based structureless filter whose unique components are shaded.
		\numcircledmod{1}-\numcircledmod{2} are described in \ref{subsec:feature-association}.
		\numcircledmod{3}-\numcircledmod{6} are presented in \ref{subsec:filter}.
		\numcircledmod{7} is discussed in \ref{subsec:state-and-error-state}.}
	\label{fig:flowchart}
\end{figure}

As consumer sensor assemblies suitable for VIO proliferate, \eg, on smartphones, 
it is desirable to calibrate the camera and the IMU in real time and without turning 
to expensive equipment. Even in cases when such special devices are available,
a VIO method with self-calibration capability is helpful,
as it can indicate the need for a new calibration.
Several methods have been proposed to tackle pose estimation and sensor calibration at
the same time.
Built upon the Multi-State Constraint Kalman Filter (MSCKF)~\cite{mourikis_multi_2007},
Li and Mourikis~\cite{li_calibration_2014} estimated a variety of parameters of a monocular camera-IMU
system, including IMU systematic errors, and camera intrinsic,
extrinsic, and temporal parameters.
Also with MSCKF, the online sensor calibration method in
\cite{eckenhoffMIMCVINS2020} estimates intrinsic and extrinsic parameters and 
time offsets of multiple cameras.
While the original MSCKF~\cite{mourikis_multi_2007} has difficulty in handling a stationary
period, its improved variant \cite{geneva_openvins_2019}
uses zero-velocity update which, however, requires delicate thresholds.

Motivated by the need for consistent covariance estimation and real-time sensor calibration,
we propose a Keyframe-based Structureless Filter (KSF) for VIO with on-the-fly
calibration capability.
A structureless estimator does not include the scene structure variables in the state vector, 
thus is typically more efficient than estimators that jointly estimate motion and scene structure.
In addition to being structureless, KSF integrates the concept of keyframes and 
sensor self-calibration.
Thanks to the use of keyframes, it can cope with standstills without resorting to zero-velocity updates.
The keyframe concept has been used in the filtering backend of \cite{sun_robust_2018},
but we use keyframes in both the feature association frontend and the filtering backend.
Besides the keyframe concept, KSF differs from
\cite{li_calibration_2014} in the concept of state variable timestamps.
Besides the two concepts, another difference to a close work \cite{eckenhoffMIMCVINS2020} is that
KSF includes old velocity variables in the state vector so that
continuous-time camera poses can be predicted with IMU measurements instead of linear fitting.

In summary, our contributions are:
\begin{itemize}
\item For efficient and consistent motion estimation with data captured
from a platform of an IMU and at least one camera, the KSF is developed, employing keyframes for image
feature association in the frontend and for state
variable management in the backend.
Its efficiency and accuracy were validated on public benchmarks, its consistency by simulation.
\item To deal with data captured by consumer sensors, the proposed
method supports calibrating IMU random biases and systematic errors,
and intrinsic, extrinsic, and temporal parameters (time delay and rolling shutter effect)
of every camera in the platform.
KSF's ability to calibrate these parameters was confirmed by
simulation and on one public benchmark captured by low-cost sensors.
\end{itemize}

Subsequently, Section \ref{sec:related} provides a non-exhaustive
overview of recent VIO methods,
focused on feature matching and sensor calibration.
Section \ref{sec:method} presents the proposed KSF covering
the state vector and error state, IMU and camera measurement models, keyframe-based feature association frontend, and the keyframe-based filtering backend.
Next, the simulation study on estimator consistency and sensor parameter
observability is reported in Section \ref{sec:simulation}.
Furthermore, Section \ref{sec:real-world-tests} gives the sensitivity analysis of various design
options, estimation accuracy of sensor parameters and poses, and
computation time on real-world data.
In the end, Section \ref{sec:discussion} discusses the results and lists caveats, 
and Section \ref{sec:conclusion} summarizes the work and points out future directions.

\section{Related Work}
\label{sec:related}
This section briefly reviews traditional geometry-based visual inertial
odometry and calibration methods without claiming to be exhaustive.
VIO methods can be loosely grouped into geometry-based, learning-based, and hybrid methods
based on the extent to which geometric models of image information are used in the
inference engine \cite{cadena2016past}. We refer to \cite{chenSurveyDeepLearning2020} for a
recent survey on learning-based and hybrid VIO methods, and 
refer to \cite{huang2019visual} for a recent survey on geometry-based VIO methods. At the time of writing this paper, 
the trending learning-based methods are reported to be robust under challenging
conditions, \eg, \cite{chen2019selective}, while geometry-based methods tend to be 
fast and accurate.
The following review focuses on the category of geometrical methods to which our proposed method belongs.

A plethora of VIO methods based on traditional geometry models has been
reported, varying in the supported number of cameras and in the calibration capability.

\subsection{Monocular Visual Inertial Odometry}
\textbf{Filtering} In \cite{strelow_motion_2004}, measurements
from a rigid camera-IMU platform were fused with a global bundle
adjustment or with an Extended Kalman Filter (EKF) to estimate platform
pose and velocity expressed in the camera frame at startup, 
landmark positions, IMU biases, and the gravity vector.
These methods were initialized with the bundle adjustment
algorithm from a special data sequence with benign motion.
With an EKF, Veth and Raquet~\cite{veth2007fusing} estimated pose and velocity of 
the sensor assembly in the navigation frame, landmark positions, IMU biases, and camera extrinsic parameters.
Feature matches were found by projecting landmarks to new images and searching in
uncertainty ellipses. Landmarks were initialized by using a terrain model or triangulation.
The MSCKF~\cite{mourikis_multi_2007} keeps a sliding
window of cloned poses for camera frames while not using landmarks in the state vector.
From every disappearing feature track (all observations corresponding to the same landmark), a landmark is triangulated,
and then its linearized observations canceling out the landmark
parameters are used to constrain multiple
cloned poses in the update step of EKF.
Also using an EKF, Jones and Soatto~\cite{jones2011visual} estimated
platform pose and velocity in an inertial frame, landmark positions, IMU biases, the gravity
vector, and camera extrinsic parameters. A landmark point is represented by its
projection direction and the logarithm of its depth.
Kelly and Sukhatme~\cite{kelly2011visual} estimated pose and velocity of the sensor
unit in a world frame (\eg, aligned to a calibration target), IMU biases, the gravity
vector, camera extrinsic parameters, and landmark positions with a UKF.
Authors of \cite{weiss_real_2012, dill_gps_optical_2015} used an EKF to fuse
IMU readings and up-to-scale motion estimates from a visual odometry method.
Li and Mourikis~\cite{li_high_2013} improved MSCKF by using first estimates of position
and velocity in computing system transition matrix and measurement Jacobians.
Hesch \etal~\cite{heschConsistencyAnalysisImprovement2014} improved MSCKF by including
distinguishable landmarks in the state vector and modifying the transition matrix and
measurement Jacobians to satisfy observability constraints.
Robocentric VIO~\cite{huaiRobocentric2018} improved MSCKF by expressing
the cloned poses relative to the first one and including the gravity vector in the state vector.

\textbf{Optimization-based Methods} With iSAM2~\cite{kaess2012isam2}, Forster \etal~\cite{forster_manifold_2017}
estimated the entire history of system states associated
to sequential keyframes. Each system
state includes pose and velocity expressed in an inertial frame and IMU biases.
The measurements encoded as factors provided to iSAM2 include a prior, preintegrated IMU factors, and
structureless projection factors constructed by eliminating the landmark parameters
from linearized reprojection errors.

With a sliding window smoother, VINS-Mono~\cite{qin_vins_2018} estimates a sliding
window of poses and velocities (expressed in a gravity-aligned world frame) and IMU
biases, inverse depth of landmarks, and extrinsic parameters and time delay of the camera.
For each selected frame, a least squares problem is constructed from image features
matched between frames by KLT~\cite{bradski2008learning}, preintegrated IMU factors,
and a marginalization factor. To bound computation, either the second latest frame or
the oldest frame is marginalized from the problem.
In \cite{he_pl_2018}, VINS-Mono was extended for optimizing both points and lines with
their observations.

\subsection{Stereo Visual Inertial Odometry}
For stereo cameras, Sun \etal~\cite{sun_robust_2018} implemented MSCKF with
observability constraints to ensure consistency. They used KLT
\cite{bradski2008learning} to track features between successive left camera images
and between left and right camera images.
In every other update step, two cloned
state variables were chosen based on relative motion from those
associated with recent frames and oldest
frames, and marginalized from the filter.

With an optimization backend, Huai \etal~\cite{huai_stereo_2015} estimated system 
state variables associated with earlier keyframes in a spatial window and recent frames
in a temporal window.
Each system state variable includes pose and velocity in a gravity-aligned world 
frame and IMU biases. The system state variables will fix their linearization points
for computing Jacobians before exiting the temporal window to ensure consistency.
Kimera-VIO~\cite{rosinolKimera2020} finds feature matches between consecutive stereo
frames by KLT~\cite{bradski2008learning}
and between left and right rectified images by correlating along
epipolar lines. By using structureless vision factors, preintegrated IMU factors,
and a prior of the system state (pose, velocity, and IMU biases), a
fixed-lag smoother optimizes the system states associated with keyframes.
The stereo VIO method in ~\cite{usenko_visual_2020} tracks features between consecutive stereo frames and 
between left and right images by an inverse-compositional sparse optical flow method.
For stereo VIO, by using reprojection errors,
preintegrated IMU factors, and a marginalization factor,
the method solves for a sliding window of system state variables associated with keyframes and
recent frames with the Gauss-Newton method.
To ensure consistency, the first-estimate Jacobian technique \cite{huangObservabilitybasedRulesDesigning2010} is used,
\ie, Jacobians of system state variables involved in a marginalization
factor are evaluated at their estimates upon marginalization.

\subsection{$N$-Camera Visual Inertial Odometry}
With the iterated EKF, ROVIO~\cite{bloesch_iterated_2017} uses the intensity
errors of image patches for landmarks to estimate robocentric pose, velocity, landmark positions, 
as well as IMU biases and camera extrinsic parameters.
The landmarks are represented by a bearing vector and an inverse depth,
allowing undelayed initialization.
With KLT~\cite{bradski2008learning} or descriptor-based matching, OpenVINS~\cite{geneva_openvins_2019} 
identifies feature matches between consecutive monocular or
stereo frames.
These matches are used by a sliding window filter to update the
state vector including pose, velocity, cloned poses of earlier frames, and landmark
parameters. It is also possible to calibrate extrinsic and intrinsic parameters of
cameras and a time delay.
For consistency, propagated values (rather than updated values) of
position, orientation, and velocity 
are used in computing the transition matrix and observation Jacobians.

Optimization-based OKVIS~\cite{leutenegger_keyframe_2015} tracks BRISK descriptors \cite{leutenegger_brisk_2011} in
a new frame of $N$ images to previous keyframes and the last frame.
With reprojection errors constructed from these feature matches, 
preintegrated IMU measurements,
a marginalization factor, and a prior of position and yaw about gravity, 
OKVIS solves for landmarks and system state variables
(each system state includes pose, velocity, and IMU biases) in a sliding window of
keyframes and recent frames.
To limit the problem size, redundant state variables of the oldest keyframe or
the old recent frame are selected and marginalized.
VINS-Fusion~\cite{qin_vins_2018} extends VINS-Mono to support stereo cameras 
and synchronous processing.

\subsection{Camera-IMU Online Calibration}
As VIO algorithms flourish, methods to calibrate a camera-IMU system have made great advances
along the lines of both offline and online calibration.
Here we skim through the offline methods and briefly review online methods.
The interested reader is referred to \cite{schneider2019observability} for recent
developments in calibration of visual inertial sensors.

Offline visual inertial calibration usually performs joint optimization of structure,
motion, and sensor parameters. Available offline approaches are achieved with varying
infrastructure support, \eg, using calibration targets
\cite{rehderExtendingKalibrCalibrating2016}, revisiting a mapped area 
\cite{lin2020infras}, or observing natural features
\cite{schneider2019observability}.

Online calibration of a camera-IMU system usually relies on filtering or fixed-lag
smoothing approaches while taking observations of existing features.
It has been common practice to estimate camera extrinsic parameters in real time, \eg,
\cite{leutenegger_keyframe_2015, jones2011visual, kelly2011visual,  bloesch_iterated_2017}.
Additional parameters have also been incorporated in real time estimators, \eg, \cite{geneva_openvins_2019}.
The extended MSCKF~\cite{li_calibration_2014} estimates scale errors, misalignment, and g-sensitivity of
the IMU and the intrinsic, extrinsic, and temporal parameters of the camera including
the frame readout time.
An improved VINS-Mono~\cite{xiao2019online} estimates scale errors,
misalignment, and g-sensitivity of the IMU.
To date, no formal observability analysis for these additional parameters exists.
But empirical evidence indicates that a rich variety of motion is necessary for
estimating these parameters reliably \cite{kelly2011visual, schneider2019observability}.
In response to the strict condition to render all parameters observable, estimators
capable of real-time calibration often come with the option to lock sensor parameters when
they are deemed unobservable or well calibrated. For a filter, 
a parameter can be locked by setting its covariance entry zero, \eg, \cite{jones2011visual}.

\section{Keyframe-based Structureless Filter}
\label{sec:method}
The problem we consider is to track the pose (position and orientation) 
and optionally calibrate the sensor parameters of a platform consisting of an IMU and 
at least one camera using inertial readings and image observations of
opportunistic landmarks with unknown positions.
Our proposed keyframe-based VIO system has two major components:
(i) the feature association frontend which processes images and gives feature matches,
and (ii) the estimation backend which fuses feature matches
and inertial measurements and estimates motion and optional calibration parameters.
Both components integrate the concept of keyframes.
In the following, we present first the notation conventions, IMU and camera measurement models, then the state variables, and finally the frontend, the backend filter and its consistency.

\subsection{Notation and Coordinate Frames}
\label{subsec:notation}
We use a $4\times 4$ transformation matrix
$\mathbf{T}_{AB}$ composed of a $3\times 3$ rotation matrix $\mathbf{R}_{AB}$
and a 3D translation vector $\mathbf{t}_{AB}$, i.e.,
$\mathbf{T}_{AB} \triangleq \left[\begin{matrix}
\mathbf{R}_{AB} & \mathbf{t}_{AB} \\
\mathbf{0}^\intercal & 1
\end{matrix}\right]$, to relate homogeneous coordinates of a point $\mathbf{p}$ in reference frame $\{A\}$, 
$\leftidx{_{A}}{\mathbf{p}} = [\leftidx{_{A}}x, \leftidx{_{A}}y, \leftidx{_{A}}z, \leftidx{_{A}}w]$, 
to those in $\{B\}$, as $\leftidx{_{A}}{\mathbf{p}} = \mathbf{T}_{AB} \cdot
\leftidx{_{B}}{\mathbf{p}}$.

This work uses several right-handed coordinate frames, 
including camera frames, IMU sensor frame $\{S\}$,
body frame $\{B\}$, and world frame $\{W\}$.
The conventional right-down-forward camera frame $\{C_k\}$ is affixed to the imaging sensor of every camera $k\in[0, N-1]$.
Without loss of generality, we designate $\{C_0\}$ as the main camera, \eg, the left camera in a stereo camera rig.
The IMU sensor frame $\{S\}$ has the origin at the intersection of the $x$- and $y$-accelerometer.
Its $x$-axis is aligned to the $x$-accelerometer, and its $y$-axis
perpendicular to the $x$-axis and close to the $y$-accelerometer 
while in the plane spanned by $x$- and $y$-accelerometer.
The estimated poses refer to a body frame $\{B\}$ affixed to the platform.
The body frame can be defined to coincide $\{S\}$ in an IMU-centric manner.
Alternatively, in a camera-centric manner, the body frame is defined to
have the same origin as $\{S\}$, and a constant orientation $R_{SnC0}$
relative to the main camera, \eg, \cite{li_calibration_2014}. 
For easy result comparison, $R_{SnC0}$ is
set to the initial value of the relative orientation between the
IMU and the main camera.
This work supports both definitions of the body frame $\{B\}$ 
(see Table~\ref{tab:imu-camera-extrinsic-models}).
For coordinate frames that evolve over time, 
we append a time parameter to such a frame to obtain a quasi-inertial
frame. 
For instance, $\{B(t)\}$ denotes the quasi-inertial frame coinciding $\{B\}$ at time $t$.

A quasi-inertial world frame $\{W\}$ is introduced for expressing the
navigation state variables of the platform.
It is instantiated at the start of a VIO estimator with accelerometers' measurements 
such that its $z$-axis is aligned with the negative gravity direction.

\subsection{IMU Measurements}
An IMU measurement model considers systematic and random errors 
in the 3-degree-of-freedom (3-DOF) gyroscope and the 3-DOF accelerometer.
Many VIO methods use a simplified IMU measurement model that only
considers random biases and noises.
A generic IMU model also accounts for systematic errors including scale factor errors,
misalignment, relative orientations, and g-sensitivity effect, \eg,
\cite{schneider2019observability, rehderExtendingKalibrCalibrating2016}.
Next, we present the simplified model first and then a generic IMU model, both being accommodated in our method.

Recall that the accelerometer triad measures the acceleration in body frame
${}_B\mathbf{a}_s$ caused by specific forces,
\begin{equation}
{}_B\mathbf{a}_s = \mathbf{R}_{WB}^\intercal \cdot (\dot{\mathbf{v}}_{WB} - \begin{bmatrix}0 & 0 & -g\end{bmatrix}^\intercal),
\end{equation}
where $g$ is the local gravity.
And the gyroscope measures angular velocity of the platform in $\{B\}$ frame ${}_B \boldsymbol{\omega}_{WB}$.

\textbf{Simple IMU model} In a simplified IMU model, the IMU measurements $\mathbf{a}_m$ and
$\boldsymbol{\omega}_m$ are assumed to be affected by accelerometer and gyroscope biases,
$\mathbf{b}_a$ and $\mathbf{b}_g$, and Gaussian white noise processes,
$\boldsymbol{\nu}_{a}$ and $\boldsymbol{\nu}_{g}$, \ie,
\begin{align}
\label{eq:simple_accel_model}
\mathbf{a}_m &= \leftidx_{B}\mathbf{a}_s + \mathbf{b}_a + \boldsymbol{\nu}_a \\
\dot{\mathbf{b}}_a &= \boldsymbol{\nu}_{ba} \\
\label{eq:simple_gyro_model}
\boldsymbol{\omega}_m &= \leftidx_{B}\boldsymbol{\omega}_{WB} +
\mathbf{b}_g + \boldsymbol{\nu}_g \\
\dot{\mathbf{b}}_g &= \boldsymbol{\nu}_{bg},
\end{align}
where the biases are assumed to be driven by Gaussian white
noise processes, $\boldsymbol{\nu}_{ba}$ and $\boldsymbol{\nu}_{bg}$.
The power spectral densities of $\boldsymbol{\nu}_{a}$, $\boldsymbol{\nu}_{g}$, 
$\boldsymbol{\nu}_{ba}$, and $\boldsymbol{\nu}_{bg}$, 
are usually assumed to be $\sigma^2_a\mathbf{I}_3$, $\sigma^2_g \mathbf{I}_3$, $\sigma^2_{ba}\mathbf{I}_3$, and $\sigma^2_{bg}\mathbf{I}_3$, respectively,
but they may have disparate diagonal values, and nonzero off-diagonal entries, \eg, for a two DOF gyro.

In summary, the IMU parameters for the simple model are
\begin{equation}
\label{eq:simple-imu}
\mathbf{x}_{imu} = \left\{\mbf b_g \enskip \mbf b_a\right\}.
\end{equation}

\textbf{Generic IMU model} Our method incorporates the generic IMU
model presented in \cite{li_calibration_2014}.
In this model, the accelerometer measurement $\mathbf{a}_m$ is corrupted by
systematic errors encoded in a $3\times 3$ matrix $\mathbf{T}_a$, 
the accelerometer bias $\mathbf{b}_a$ and noise process $\boldsymbol{\nu}_a$,
\begin{equation}
\label{eq:accel_model}
\begin{split}
\mathbf{a}_m &= \mathbf{T}_a {}_B \mathbf{a}_s + \mathbf{b}_a + \boldsymbol{\nu}_a \\
\dot{\mathbf{b}}_a &= \boldsymbol{\nu}_{ba}.
\end{split}
\end{equation}
The 9 entries of $\mathbf{T}_a$ encompasses 3-DOF scale factor error, 3-DOF misalignment,
and 3-DOF relative orientation between the accelerometer input axes and $\{B\}$ frame.

The gyroscope measurement $\boldsymbol{\omega}_m$ is corrupted by systematic errors encoded in a 
$3\times 3$ matrix $\mathbf{T}_g$ and g-sensitivity effect encoded in a $3\times 3$ matrix $\mathbf{T}_s$,
the gyroscope bias $\mathbf{b}_g$ and noise process $\boldsymbol{\nu}_g$,
\begin{equation}
\label{eq:gyro_model}
\begin{split}
\boldsymbol{\omega}_m &= \mathbf{T}_g {}_B \boldsymbol{\omega}_{WB} +
\mathbf{T}_s {}_B \mathbf{a}_s + 
\mathbf{b}_g + \boldsymbol{\nu}_g \\
\dot{\mathbf{b}}_g &= \boldsymbol{\nu}_{bg}.
\end{split}
\end{equation}
The 9 entries of $\mathbf{T}_\omega$ encompasses 3-DOF scale factor error, 3-DOF misalignment,
and 3-DOF relative orientation between the gyroscope input axes and $\{B\}$ frame.

Alternatively, the IMU measurements can be modeled with separate coefficients 
for scaling, misalignment, and bearing effects as done in
\cite{rehderExtendingKalibrCalibrating2016}, thus it will be easier to interpret the
resulting estimates.
But to avoid dealing with SO(3) relative orientations in implementation,
we adopt the above formulation, \eqref{eq:accel_model} and \eqref{eq:gyro_model}.
In this case, the IMU parameters $\mbf x_{imu}$ are summarily,
\begin{equation}
\label{eq:IMU-parameters}
\mbf x_{imu} = \left\{\mbf b_g \enskip \mbf b_a
\enskip \overset{\rightarrow}{\mbf{T}}_g \enskip \overset{\rightarrow}{\mbf{T}}_s 
\enskip \overset{\rightarrow}{\mbf{T}}_a \right\}
\end{equation}
where $\overset{\rightarrow}{(\cdot)}$ stacks entries of a matrix into a vector.

For a well-constrained system, the presented IMU models and compatible
camera extrinsic models are listed in Table~\ref{tab:imu-camera-extrinsic-models}.
For the generic IMU model \eqref{eq:IMU-parameters},
it is necessary to use the body frame defined in a camera-centric
manner (Section \ref{subsec:notation}).
Otherwise, if the IMU-centric $\{B\}$ frame is used,
$\mathbf{R}_{BC0}$ shall be unknown and thus variable,
making the 3-DOF relative orientation in $\mathbf{T}_a$ redundant.
On the other hand, for the simple IMU model \eqref{eq:simple-imu}, it
is desirable to use the body frame defined in an IMU-centric manner for
the purpose of estimating the relative orientation $\mathbf{R}_{BC0}$.

\begin{table}[]
	\centering
	\caption{Compatible IMU and camera extrinsic parameters.}
	\label{tab:imu-camera-extrinsic-models}
	\begin{tabular}{llll}
		\hline
		\begin{tabular}[c]{@{}l@{}}Body frame\\ definition\end{tabular} &
		\begin{tabular}[c]{@{}l@{}}IMU \\ parameters\end{tabular} &
		\begin{tabular}[c]{@{}l@{}}Main camera \\ extrinsics \end{tabular} &
		\begin{tabular}[c]{@{}l@{}}Other camera \\ extrinsics\end{tabular} \\ \hline
		\begin{tabular}[c]{@{}l@{}}Camera-centric\\ body frame\end{tabular} &
		\begin{tabular}[c]{@{}l@{}}$\mathbf{b}_g,  \mathbf{b}_a$\\ $\mathbf{T}_g, \mathbf{T}_s,\mathbf{T}_a$\end{tabular} &
		\begin{tabular}[c]{@{}l@{}}$\mathbf{R}_{C0B}=\mathrm{const}$\\
			$\mathbf{t}_{C0B}$
		\end{tabular} &
		\begin{tabular}[c]{@{}l@{}}$\mathbf{T}_{BCk}$ or\\ $\mathbf{T}_{C0Ck}$\end{tabular} \\ \hline
		\begin{tabular}[c]{@{}l@{}}IMU-centric\\body frame\end{tabular} &
		$\mathbf{b}_g,  \mathbf{b}_a$ &
		$\mathbf{T}_{BC}$ &
		\begin{tabular}[c]{@{}l@{}}$\mathbf{T}_{BCk}$ or\\ $\mathbf{T}_{C0Ck}$\end{tabular} \\ \hline
	\end{tabular}
\end{table}

\textbf{Propagation with IMU Data}
With a sequence of IMU readings~$\mathbf{z}_{imu, k}$, between $t_{k-1}$ and $t_k$, 
the navigation state variable $\mbc{\pi}(t_{k})$, and the covariance
of $\mbc{\pi}(t_{k})$ and IMU parameters $\mathbf{x}_{imu}(t_{k})$,
$\boldsymbol{\Sigma}(\mbc{\pi}(t_{k}), \mathbf{x}_{imu}(t_{k}))$, can be propagated from
given state variables $\mbc{\pi}(t_{k-1})$, $\mbf{x}_{imu}(t_{k-1})$, 
and covariance $\boldsymbol{\Sigma}(\mbc{\pi}(t_{k-1}), \mbf{x}_{imu}(t_{k-1}))$ at $t_{k-1}$ 
as expressed by $\mathbf{f}(\cdot)$,
\begin{equation}\label{eq:imu_prop}
\begin{split}
\left\{\mbc{\pi}(t_k), \boldsymbol{\Sigma}(\mbc{\pi}(t_k),\mbf{x}_{imu}(t_{k})), \boldsymbol{\Phi}(t_k, t_{k-1}) \right\} 
&= \\
\mathbf{f}(\mbc{\pi}(t_{k-1}), \mbf{x}_{imu}(t_{k-1}),
\boldsymbol{\Sigma}(\mbc{\pi}(t_{k-1}),
\mbf{x}_{imu}(t_{k-1}))&, \\
\mathbf{z}_{imu, k}, \mbf n_{imu})&,
\end{split}
\end{equation}
where the noises of IMU readings are stacked in $\mbf n_{imu}$, \ie,
\begin{equation}
\mbf n_{imu} = [\boldsymbol{\nu}_g^\intercal, \boldsymbol{\nu}_a^\intercal, 
\boldsymbol{\nu}_{bg}^\intercal, \boldsymbol{\nu}_{ba}^\intercal]^\intercal.
\end{equation}

The transition matrix $\boldsymbol{\Phi}(t_k, t_{k-1})$ is the derivative of the propagated error state,
$\delta \mbc{\pi}(t_{k})$ and $\delta \mbf{x}_{imu}(t_{k})$, relative to
the initial error state, $\delta \mbc{\pi}(t_{k-1})$ and $\delta \mbf{x}_{imu}(t_{k-1})$, \ie,
\begin{equation}
\label{eq:process_jac}
\boldsymbol{\Phi}(t_k, t_{k-1}) = \frac{\partial (\delta \mbc{\pi}(t_{k}), \delta \mbf{x}_{imu}(t_{k}))}
{\partial (\delta \mbc{\pi}(t_{k-1}), \delta \mbf{x}_{imu}(t_{k-1}))}.
\end{equation}
Detailed calculation of $\boldsymbol{\Phi}(t_k, t_{k-1})$ and $\boldsymbol{\Sigma}(\mbc{\pi}(t_k),\mbf{x}_{imu}(t_{k}))$ in
\eqref{eq:imu_prop} can be found in \cite[Appendix A]{huai_collaborative_2017}, \cite{jekeli_inertial_2001}.
The propagation \eqref{eq:imu_prop} should support backward integration in time to deal with observations earlier than a pose variable.

\subsection{Camera Measurements}
The visual measurements are expressed by conventional reprojection errors.
Our adopted pinhole camera observation model relates a landmark
$\mathbf{L}_i$'s observation $\mathbf{z}_{i,j}^k$ in
image $k$ of frame $j$ at time $t_{i,j}^k$ to its parameters and the system state.
For a landmark $\mathbf{L}_i$ expressed in $\{W\}$,
$\leftidx{_W}{\mathbf{p}}_i = [x_i, y_i, z_i, 1]^\intercal$,
the observation model is given by
\begin{equation}
\label{eq:reprojection}
\begin{split}
\mathbf{z}_{i, j}^k &= \begin{bmatrix}
u & v \end{bmatrix}^\intercal \\
&= \mathbf{h}^k(
\mathbf{T}_{BC_k}^{-1} \mathbf{T}_{WB(t_{i,j}^k)}^{-1} 
\leftidx{_W}{\mathbf{p}}_{i}) + \mathbf{w}_c
\end{split}
\end{equation}
where the perspective projection function $\mathbf{h}^k(\cdot)$
factors in camera intrinsic parameters, $\mathbf{x}_c^k$, 
and we assume that an image observation is affected by
Gaussian noise $\mathbf{w}_c \sim N(\mathbf{0}, \sigma_c^2\mathbf{I}_2)$.

Alternatively, a landmark $\mathbf{L}_i$ may be anchored 
in camera $b$ of frame $a$, $\{C_b(t_a)\}$,
where $t_a$ is frame $a$'s timestamp.
Then, $\mathbf{L}_i$ can be represented by a homogeneous point in terms of its inverse depth
$\rho_i$, and its observation direction $[\alpha_i, \beta_i, 1]$, \ie,
$\leftidx{_{C_b(t_a)}}{\mathbf{p}}_i = \frac{1}{\rho_i}[\alpha_i, \beta_i, 1, \rho_i]^\intercal$,
and the observation model is given by
\begin{equation}
\label{eq:reprojection_anchor}
\begin{split}
\mathbf{z}_{i, j}^k
&= \mathbf{h}^k(
\mathbf{T}_{BCk}^{-1} \mathbf{T}_{WB(t_{i,j}^k)}^{-1} 
\mathbf{T}_{WB(t_a)} \mathbf{T}_{BC_{b}} \cdot \\
&\qquad \quad \leftidx{_{C_b(t_a)}}{\mathbf{p}}_{i}) + \mathbf{w}_c
\end{split}
\end{equation}

Notice that the enclosed term in \eqref{eq:reprojection} or \eqref{eq:reprojection_anchor} is 
obtained by transforming a point (\ie, $\mathbf{Tp}$) or inversely transforming a point (\ie, $\mathbf{T^{-1}p}$),
their Jacobians can be computed in a unified manner by recursively using
the chain rule as done in the reverse-mode automatic differentiation \cite{geronHandsonMachineLearning2017}.
Because in general the camera measurements in one image are at different epochs
due to the rolling shutter effect, the Jacobians of the pose at feature observation
epoch $t_{i,j}^k$, $\mathbf{T}_{WB(t_{i,j}^k)}$, relative to the pose and velocity at $t_j$, $\mathbf{T}_{WB(t_j)}$ and $\mathbf{v}_{WB(t_j)}$,
are computed
with IMU propagation matrix $\boldsymbol{\Phi}$ \eqref{eq:process_jac}.
Jacobians relative to time parameters, \eg, camera time delay,
are computed from the velocity of the propagated pose $\mathbf{T}_{WB(t_{i,j}^k)}$.
The measurement Jacobian relative to IMU parameters is ignored as in
\cite{li_vision_2014} because its entries are usually small.

\subsection{State and Error State}
\label{subsec:state-and-error-state}
This section presents variables used in formulating the nonlinear VIO problem.
We consider the true state $\mathbf{x}$ and the associated
error state $\delta \mathbf{x}$ to be unknown random variables,
and denote the state estimate (mean) by $\hat{\mathbf{x}}$.

\textbf{State Vector}
The state of a VIO system $\mathbf{x}(t)$ consists of current navigation state variable $\boldsymbol{\pi}(t)$,
IMU parameters $\mathbf{x}_{imu}$, parameters of $N$ cameras $\mathbf{x}_c$,
and a sliding window of past navigation state variables $\mathbf{x}_w$
at known epochs $\{t_j\}$
where $j$ enumerates past navigation state variables for $N_{kf}$ keyframes and
$N_{tf}$ recent frames in the sliding window.
That is,
\begin{align}
\label{eq:all_states} 
\mathbf{x}(t) &= \{\boldsymbol{\pi}(t), \mathbf{x}_{imu}, \mathbf{x}_c,
\mathbf{x}_w\}\\
\label{eq:ncamera_params}
\mathbf{x}_c &= \{\mathbf{x}_{c}^k | k = 0, 1, \cdots, N-1\} \\
\label{eq:slidingwindow_states}
\mathbf{x}_w &= \{\boldsymbol{\pi}(t_j) | j = 0, 1, \cdots N_{kf} + N_{tf} -1 \}.
\end{align}

The navigation state $\boldsymbol{\pi}(t)$ expressed in $\{W\}$ includes position
$\mathbf{t}_{WB(t)}$, orientation $\mathbf{R}_{WB(t)}$, and velocity $\mathbf{v}_{WB(t)}$ of the platform at time $t$, \ie,
\begin{equation}
\label{eq:nav_states}
\boldsymbol{\pi}(t) := \{\mathbf{t}_{WB(t)}, \mathbf{R}_{WB(t)}, \mathbf{v}_{WB(t)}\}.
\end{equation}

The IMU parameters $\mathbf{x}_{imu}$ are given in \eqref{eq:simple-imu} or \eqref{eq:IMU-parameters}.

\textbf{Camera Parameters} The parameters for one camera $\mathbf{x}_c^k$ include 
the camera extrinsic,
intrinsic and temporal parameters.
The extrinsic parameters of the main camera in the $N$-camera system
are represented by $\mathbf{T}_{BC0}$.
For other cameras, the extrinsic variable can be parameterized by either
$\mathbf{T}_{BCk}$ or $\mathbf{T}_{BC0} \cdot \mathbf{T}_{C0Ck}$.
The intrinsic parameters of one camera include focal length, principal point, and
distortion parameters when assuming a pinhole projection model.
The temporal parameters of camera $k$ include the relative time delay
$t_{d}^k$ between camera $k$ and the IMU, and frame readout time $t_{r}^k$.
Without loss of generality, $t_{d}^k$ is defined such that the raw timestamp of an image $k$ of frame
$j$ by the camera clock, $t_{c,j}^k$, plus $t_{d}^k$, is the timestamp by the IMU clock
for the central row or mid-exposure of the image, $t_j^k$, \ie,
$t_j^k = t_{c,j}^k + t_{d}^k$.

Each navigation state in the sliding window, $\boldsymbol{\pi}(t_j)$, associated with frame $j$ 
is at a known epoch $t_j$ which is set to the first estimate
of the IMU timestamp for the central row of the main image of frame $j$, \ie,
\begin{equation}
\label{eq:state_epoch}
t_j = t_{c,j}^0 + t_{d,j}^0,
\end{equation}
where $t_{d,j}^0$ is the estimate of the main camera's time delay 
$t_{d}^0$ at processing frame $j$ and hence known.
Tying navigation state variables to known epochs is conceptually simpler than tying them to true epochs of estimated values
(\cite{geneva_openvins_2019, li_vision_2014}).

With time delay $t_{d}^k$, and readout time $t_{r}^k$ for camera $k$,
the epoch to observe a landmark $i$ at $[u, v]$ in 
image $k$ of frame $j$ with raw timestamp $t_{c, j}^k$ and $h$ rows is computed by
\begin{align}
\label{eq:feature_time}
t_{i, j}^k = t_{c,j}^k + t_{d}^k + (\frac{v - 0.5h}{h}) t_{r}^k
\end{align}
Since landmarks are observed at varying epochs due to the rolling shutter effect,
velocity variables are included in the sliding window \eqref{eq:slidingwindow_states} so that
the camera pose for a landmark observation can be
propagated from the relevant navigation state variable with IMU measurements.

\textbf{Random Constant State Variables}
Among all state variables, only the current navigation state variable
$\boldsymbol{\pi}(t)$ and the IMU biases are time-variant.
The historic navigation state variables $\mathbf{x}_w$ are unknown random constants \cite{jekeli_inertial_2001} 
because they are at known epochs.
The IMU parameters $\mathbf{x}_{imu}$ \eqref{eq:simple-imu} or \eqref{eq:IMU-parameters} excluding biases and 
the camera parameters $\mathbf{x}_c$ are assumed to be random constants \cite{jekeli_inertial_2001}
because these parameters change little within the timespan of the sliding window.
This practice was adopted in 
\cite{li_calibration_2014,guoEfficientVisualInertialNavigation2014} and
OpenVINS~\cite{geneva_openvins_2019}.

If a sensor parameter is in fact not constant over time, 
noise can be added to account for its additional uncertainty as is done for IMU biases.
This technique has been used for estimating time delay between a
camera and an IMU in \cite[(32)]{liOnlineTemporalCalibration2014}.

The motivation for including these sensor parameters in an estimator is threefold.
First, the hardware is usually imperfect, especially for sensors on mobile devices \cite{huai_mobile_2019}.
Secondly, estimating such random constant parameters incurs marginal computation.
Thirdly, these parameters can be locked up online and thus excluded from estimation ad hoc.
For a filter, this is done by zeroing out their covariance blocks.

\textbf{Error State Vector}
All state variables fit in the $\boxplus$-manifolds axiomatized in
\cite{hertzberg_integrating_2011}. Therefore, the error $\delta\mathbf{x}$
for a state variable $\mathbf{x}$ on a $\boxplus$-manifold is defined
with the $\boxplus$-operator associated with the manifold,
\begin{equation}
\mathbf{x} = \hat{\mathbf{x}} \boxplus \delta\mathbf{x}.
\label{eq:boxplus}
\end{equation}
In this sense, $\boxplus$ generalizes the conventional $+$.
The error state variables are mostly defined with the $+$-operator, \eg, 
for translation $\mathbf{t}_{WB}$, velocity $\mathbf{v}_{WB}$, and gyroscope bias $\mathbf{b}_g$,
\begin{subequations}
	\begin{align}
	\label{eq:delta_t_WB}
	\mathbf{t}_{WB} &= \hat{\mathbf{t}}_{WB} + \delta \mathbf{t}_{WB} \\
	\label{eq:delta_v_WB}
	\mathbf{v}_{WB} &= \hat{\mathbf{v}}_{WB} + \delta \mathbf{v}_{WB} \\
	\label{eq:delta_bias_omega}
	\mathbf{b}_g &= \hat{\mathbf{b}}_g + \delta \mathbf{b}_g.
	\end{align}
\end{subequations}

The exceptions are rotations.
For an element $\mathbf{R}_{WB}$ in the space of orientation preserving rotations SO(3),
its error state $\delta \boldsymbol{\theta}_{WB}$ is defined such that
right multiplying the matrix exponential of $\delta \boldsymbol{\theta}_{WB}^\times$
by the estimate $\hat{\mathbf{R}}_{WB}$ gives $\mathbf{R}_{WB}$, \ie,
\begin{equation}
\label{eq:theta_WB}
\mathbf{R}_{WB} = \exp(\delta \boldsymbol{\theta}_{WB}^\times) \hat{\mathbf{R}}_{WB}
 \approx (\mathbf{I} + \delta \boldsymbol{\theta}_{WB}^\times) \cdot \hat{\mathbf{R}}_{WB},
\end{equation}
where $(\cdot)^\times$ obtains the skew-symmetric matrix for a 3D vector.
The small angle approximation in the last expression is useful in deriving
analytical Jacobians.
Equivalently, in terms of Hamilton quaternions,
\begin{equation}
	\mathbf{q}_{WB} = \begin{bmatrix}
	\frac{\delta\boldsymbol{\theta}_{WB}}{\Vert\delta\boldsymbol{\theta}_{WB}\Vert}
	\sin\frac{\Vert\delta\boldsymbol{\theta}_{WB}\Vert}{2} \\ \cos(\Vert\delta\boldsymbol{\theta}_{WB}\Vert/2)
	\end{bmatrix} \otimes \hat{\mathbf{q}}_{WB}
	\approx \begin{bmatrix}
	\frac{\delta\boldsymbol{\theta}_{WB}}{2} \\ 1
	\end{bmatrix} \otimes \hat{\mathbf{q}}_{WB},
\end{equation}
where $\mathbf{q}_{WB}$ and $\hat{\mathbf{q}}_{WB}$ are the quaternions for 
$\mathbf{R}_{WB}$ and $\hat{\mathbf{R}}_{WB}$, respectively, 
and $\otimes$ is the quaternion multiplication operator \cite[(12)]{solaQuaternionKinematicsErrorstate2017}.

\textbf{Camera Extrinsic Error Variables}
The error variable of the main camera's extrinsic parameters is defined
considering the definition of the body frame $\{B\}$ (Section
\ref{subsec:notation}). And the error variables of other cameras'
extrinsic parameters are defined considering specific parameterizations.

For the main camera, if the camera-centric body frame is
used, then $\mathbf{T}_{BC0}$ is fixed in orientation,
and the extrinsic error is only defined in the translation component 
which can be expressed by either $\mathbf{t}_{BC0}$ or $\mathbf{t}_{C0B}$.
Following the notation of \cite{li_calibration_2014}, the latter is used to define the extrinsic error, \ie,
\begin{equation}
	\mathbf{t}_{C0B} = \hat{\mathbf{t}}_{C0B} + \delta \mathbf{t}_{C0B}.
\end{equation}

Otherwise, when the IMU-centric definition of $\{B\}$ is adopted, the main camera's
extrinsic error consists of a rotation part $\delta\boldsymbol{\theta}_{BC0}$ and a translation part $\delta \mathbf{t}_{BC0}$, \ie,
\begin{align}
\mathbf{t}_{BC0} &= \hat{\mathbf{t}}_{BC0} + \delta \mathbf{t}_{BC0} \\
\mathbf{R}_{BC0} &= \exp(\delta \boldsymbol{\theta}_{BC0}^\times) \hat{\mathbf{R}}_{BC0}
\end{align}

For other cameras, their extrinsic parameters, $\mathbf{T}_{BCk}$, can be
parameterized relative to the body frame, \ie, by themselves,
\begin{equation}
\mathbf{T}_{BCk} = \mathbf{T}_{BCk} \enskip k \in [1, N-1],
\end{equation}
or to the main camera, \ie,
\begin{equation}
	\mathbf{T}_{BCk} = \mathbf{T}_{BC0} \cdot \mathbf{T}_{C0Ck} \enskip k \in [1, N-1].
\end{equation}
The error variables of $\mathbf{T}_{BCk}$ and $\mathbf{T}_{C0Ck}$
are defined for the translation and orientation components
like \eqref{eq:delta_t_WB} and \eqref{eq:theta_WB}, \ie, for $k \in [1, N-1]$,
\begin{subequations}
\begin{align}
	\mathbf{t}_{BCk} &= \hat{\mathbf{t}}_{BCk} + \delta \mathbf{t}_{BCk} \\
	\mathbf{R}_{BCk} &= \exp(\delta \boldsymbol{\theta}_{BCk}^\times) \hat{\mathbf{R}}_{BCk} \\
	\mathbf{t}_{C0Ck} &= \hat{\mathbf{t}}_{C0Ck} + \delta \mathbf{t}_{C0Ck} \\
	\mathbf{R}_{C0Ck} &= \exp(\delta \boldsymbol{\theta}_{C0Ck}^\times) \hat{\mathbf{R}}_{C0Ck}.
\end{align}
\end{subequations}

\subsection{Feature Association Frontend}
\label{subsec:feature-association}
The frontend extracts and matches image features by using BRISK descriptors \cite{leutenegger_brisk_2011} and 
removes outliers by RANSAC methods provided in OpenGV~\cite{kneip_OpenGV_2014},
similarly to \cite{leutenegger_keyframe_2015}.

\begin{figure}[!t]
	\centering
	\includegraphics[width=\columnwidth]{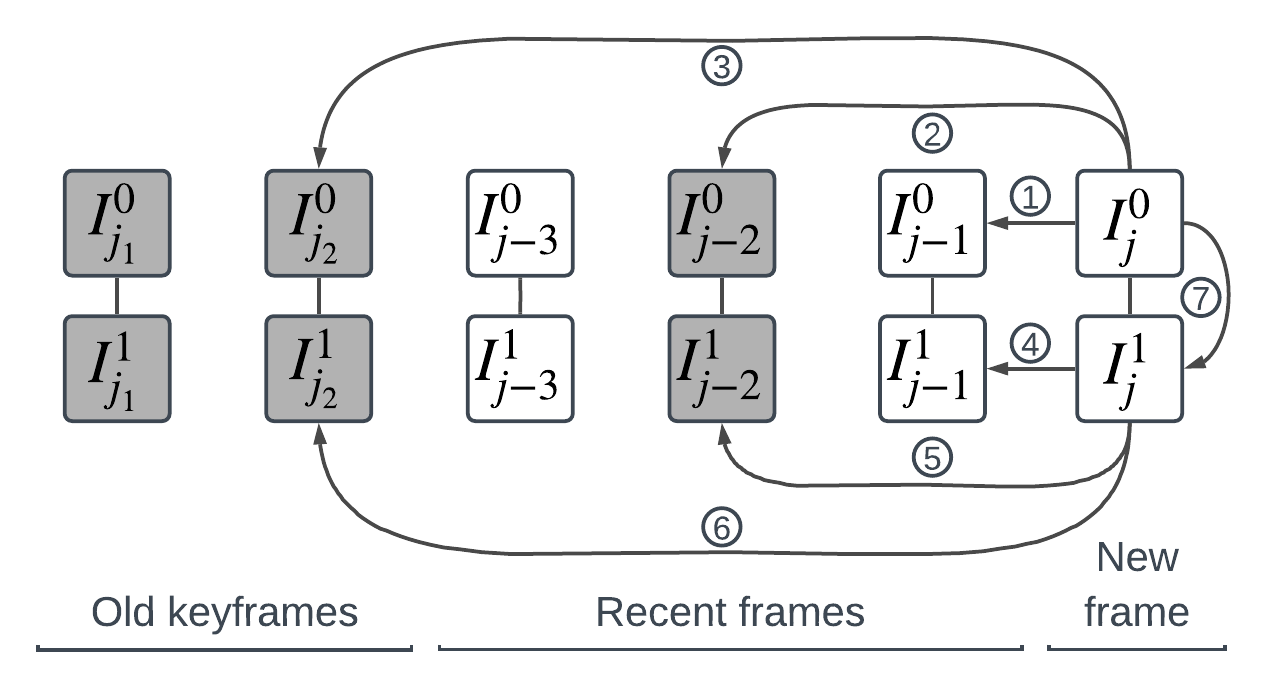}
	\caption{The keyframe-based feature matching consists of matching the current
		frame to the last frame, \numcircledmod{1} and \numcircledmod{4}, matching the
		current frame to two previous keyframes, \numcircledmod{2}, \numcircledmod{3},
		\numcircledmod{5}, and \numcircledmod{6}, and matching between images of the
		current frame, \numcircledmod{7}. Showing keyframes as shaded frames, this
		diagram assumes a platform with stereo cameras.
		\numcircledmod{1}-\numcircledmod{6} each includes a 3D-2D matching phase and a 2D-2D matching phase (Section \ref{subsec:feature-association}).
		For a single camera setup, only \numcircledmod{1}-\numcircledmod{3} are needed.
	}
	\label{fig:feature-matching}
\end{figure}

Consider a frame captured by $N$ synchronized cameras.
Each image in the frame is taken by one camera, and these images are grouped into 
a frame because their timestamps are close (but not necessarily identical).
The matching procedure associates features of the current frame to several previous frames which 
include the last frame and two most recent keyframes, and associates
features between images of the current frame if applicable (Fig.~\ref{fig:feature-matching}).

For every previous frame, the matching operation consists of a 3D-2D
matching phase for features associated with 3D landmarks and 
a 2D-2D matching phase for features not yet triangulated,
both done by brute-force matching.
Though a bit counter intuitive, brute-force matching is quite efficient
with up to 400 features per image (see Section \ref{subsec:timing}).

Given two frames, for every pair of images captured by the same camera, 
the 3D-2D phase matches descriptors of triangulated 3D landmarks observed in the earlier image to descriptors in the latter image.
Each match is then checked by projecting the 3D landmark onto the
latter image. Lastly, all valid matches go through a three-point
absolute pose RANSAC step \cite{kneip_OpenGV_2014} to further remove outliers.
Given two frames, for every pair of images captured by the same camera, 
the 2D-2D phase matches descriptors not associated with landmarks in the earlier
image to those in the latter image.
Then from these feature matches, new landmarks will be triangulated.
Notably, this triangulation step is not required by the backend estimator.
Its purpose is to remove feature matches of large projection errors while keeping
those of low rotation-compensated disparity, and to provide depth estimates for 
3D-2D matching.
After successful triangulation, these 2D-2D matches will go through a two-point
relative pose RANSAC step \cite{kneip_robust_2011} to remove outliers.
Each valid 2D-2D match spawns a feature track which is all observations of the triangulated landmark.
These feature tracks are shared with the backend estimator.
On the other hand, they are to be extended in future 3D-2D and inter-camera matching steps.

In the case of $N$ cameras indexed by $k\in [0, N-1]$, features are associated between images captured by
cameras of overlapping field-of-views, \eg, $\{(0, 1), (1, 2), \cdots, (N-2, N-1), (N-1, 0)\}$.
Unlike \cite{leutenegger_keyframe_2015}, matches that deviate much from
the epipolar lines will be removed to deal with outliers.
Also, a match of two features of two landmarks will cause the two landmarks and their
feature tracks to be fused.

The above matching procedure distinguishes ordinary frames and keyframes.
Keyframes are selected by using the criteria in \cite{leutenegger_keyframe_2015}.
For every image $k$ in the current frame, we compute the overlap $o_k$ between
the convex hull $H_k$ of 2D image features associated with landmarks
and the one of all 2D image features, and the ratio $r_k$ between the
number of 2D image features associated with landmarks
and the number of those fall into the convex hull $H_k$.
A keyframe is selected if the maximum of area overlaps $\{o_k| k \in [0, N-1]\}$ is
less than a threshold $T_o$ (typically 60\%) or 
if the maximum of match ratios $\{r_k| k \in [0, N-1]\}$ is less than $T_r$ (typically 20\%).

\subsection{Filter Workflow}
\label{subsec:filter}
The backend, a keyframe-based structureless filter, estimates state
variables tied to frames and keyframes using completed feature tracks and 
marginalizes old or redundant state variables (Fig.~\ref{fig:flowchart}).
The filter is structureless since it does not include landmarks in the state
vector, similarly to MSCKF~\cite{li_high_2013}.
The novelty of the filter is that it manages motion-related state variables depending on whether 
they are associated with ordinary frames or keyframes.
The keyframe-based state management scheme is facilitated by the selection of keyframes in the frontend and 
is motivated by the need to make a structureless filter work when the sensor setup is stationary.

At the beginning (Algorithm~\ref{algo:MSCKF}), the VIO system is
initialized to zero position, zero velocity, zero accelerometer bias, 
gyroscope bias averaged over first few gyroscope readings, and an orientation 
such that the $z$-axis of the world frame $\{W\}$ is aligned to the negative
gravity direction by using a few accelerometer readings.
The other IMU parameters \eqref{eq:IMU-parameters}, 
the camera extrinsic parameters $\mathbf{T}_{BCk}$, time delay $t_d^k$,
and readout time $t_r^k$, are initialized to their nominal values from datasheets or experience.

The standard deviations of state variables \eqref{eq:ncamera_params}-\eqref{eq:nav_states}
are usually initialized to sensible values, see Table~\ref{tab:init_value_std} 
for our simulation setup and Table~\ref{tab:tumvi-room-calib}
for real data tests.

\begin{figure}[]
\centering
\includegraphics[width=0.95\columnwidth]{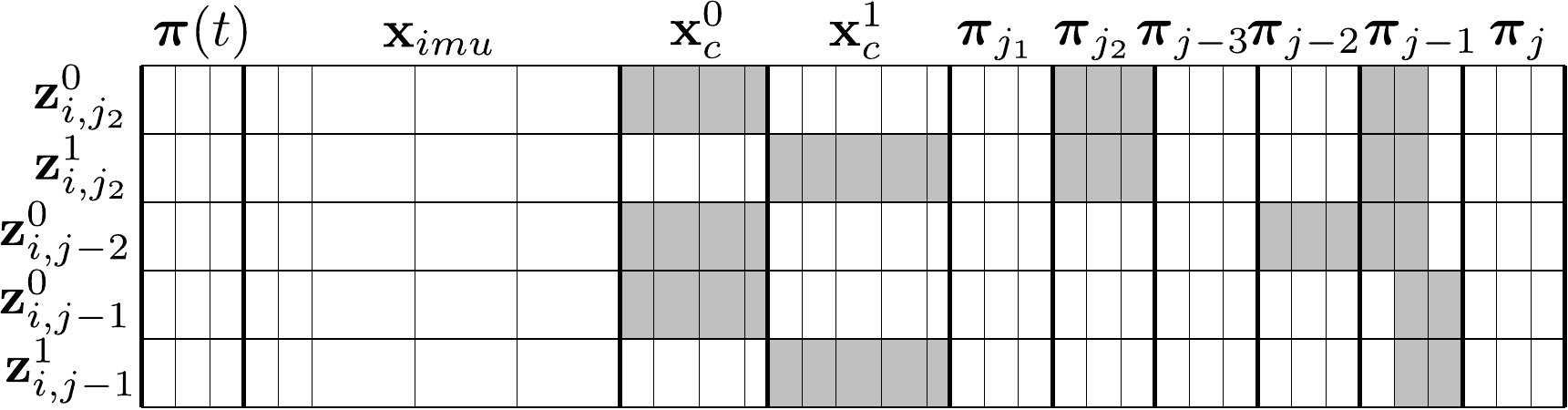}
\caption{Schematic drawing of a landmark $i$'s observation Jacobians relative to
the state vector whose nonzero entries are shaded. In accordance with
Fig.~\ref{fig:feature-matching}, as a new frame $j$ arrives, this landmark
completes its feature track at frame $j-1$ which serves as its anchor frame
\eqref{eq:reprojection_anchor}. After the navigation state variable for frame
$j$ has been cloned as $\boldsymbol{\pi}_j:=\boldsymbol{\pi}(t_j)$, the computed
Jacobians for $i$ are stacked as shown and ready for canceling out the
Jacobian for $i$'s parameters (Section \ref{subsec:filter}).}
\label{fig:jacobians}
\end{figure}

As a frame of coarsely synchronized $N$ images arrives, the navigation state \eqref{eq:nav_states} and 
the covariance matrix are propagated with inertial readings to the state epoch 
(\eqref{eq:state_epoch}, Box~\numcircledmod{3} of Fig.~\ref{fig:flowchart}) for the frame.
Then, the navigation state variable is cloned and augmented to the state vector and the
covariance matrix is also expanded for the cloned navigation state
(Line~\ref{MSCKF:line:propagate}, Box~\numcircledmod{4} of Fig.~\ref{fig:flowchart}).
Because the cloned navigation state is assigned a known constant epoch \eqref{eq:state_epoch},
unlike \cite{geneva_openvins_2019} or \cite[(30)]{li_vision_2014}, we
need not to account for uncertainty of the time delay estimate when expanding
covariance for the cloned state. But we need to compute the measurement Jacobian
relative to time delays of cameras in the update step (Line~\ref{MSCKF:line:update}).
In parallel, feature descriptors are extracted from images in the frame and
matched to those in the last frame and two recent keyframes 
(Line~\ref{MSCKF:line:descriptor},
Boxes \numcircledmod{1} and \numcircledmod{2} of
Fig.~\ref{fig:flowchart}, and Fig.~\ref{fig:feature-matching}) as in
the frontend of \cite{leutenegger_keyframe_2015}. Feature tracks
that disappear in the current frame will be used as measurements for update.

A completed feature track with at least three observations is used to
triangulate a point landmark and all observations in the track provide
reprojection errors to the filter
(Lines~\ref{MSCKF:line:triangulate}-\ref{MSCKF:line:canceljac}). 
A completed feature track of two observations are discarded because they are not
informative to the filter \cite{sun_robust_2018}.
For an example landmark $i$, the Jacobians of its observations relative to the
state vector are visualized in Fig.~\ref{fig:jacobians}.
Since the landmark is not represented in the state vector, its Jacobian is
canceled out by projecting all measurement Jacobians to the left nullspace of
the landmark's Jacobian \cite{mourikis_multi_2007}.
The left nullspace is typically spanned by $2n-3$ column vectors for a
landmark well observed $n$ times \cite[4.51]{huai_collaborative_2017},
but has $2n-2$ columns for a landmark at infinity because of the rank-deficient landmark Jacobian.
This special case is considered in canceling out the landmark Jacobian and 
in performing the subsequent Mahalanobis gating test.
To remove outliers of large measurement residuals, 
every reprojection measurement goes
through a Mahalanobis gating test (Line~\ref{MSCKF:line:mahal}) \cite{li_high_2013}.
All valid residuals and Jacobians for these disappeared landmarks are stacked
and used to update the state vector and the covariance matrix 
(Line~\ref{MSCKF:line:update}, Box~\numcircledmod{5} of Fig.~\ref{fig:flowchart})
\cite{mourikis_multi_2007}.

To bound computation, redundant frames are selected and marginalized from
the filter (Lines~\ref{MSCKF:line:detect-dud-frames}-\ref{MSCKF:line:remove-dud-variables}, Box~\numcircledmod{6} of
Fig.~\ref{fig:flowchart}) once the number of navigation
state variables in the sliding window exceeds the total number of allowed old
keyframes $N_{kf}$ and recent frames $N_{tf}$ successive in time, 
$N_{kf} + N_{tf}$.
In each marginalization operation, at least $X$ redundant frames 
($X$ is 3 for the monocular case, 2 for multiple camera setups) are
chosen (Line~\ref{MSCKF:line:detect-dud-frames}) because two reprojection
measurements for a unknown landmark is
uninformative \cite{sun_robust_2018}. To meet this requirement, they are chosen
first among the recent non-keyframes while excluding the most recent $N_{tf}$ frames
and secondly among the oldest keyframes. For the case of
Fig.~\ref{fig:feature-matching}, if $N_{kf}=2$ and $N_{tf}=3$, then the
redundant frames are keyframe $j_1$ and frame $j-3$ for the latest frame $j$.

With these redundant frames, we update the filter with observations of landmarks
each observed more than twice in such frames
(Line~\ref{MSCKF:line:update-with-dud-frames}).
For such a landmark, if it can be triangulated with its entire observation history, its
observations in the redundant frames are used for Kalman update. All other observations in the
redundant frames are discarded. 
After the update, corresponding state variables and
entries in the covariance matrix for these redundant frames are removed
(Line~\ref{MSCKF:line:remove-dud-variables}).

\begin{algorithm}
\caption{Keyframe-based structureless filter workflow.
Changes relative to MSCKF~\cite{mourikis_multi_2007} are shown in \textit{italics}.}
	\label{algo:MSCKF}
	\begin{algorithmic}[1]
		\State Initialize state and covariance
		\label{MSCKF:line:initialize}
		\While{a frame of $N$ images arrives}
		\State Propagate and augment state vector and covariance
		\label{MSCKF:line:propagate}
		\State \textit{Extract descriptors and match to last frame and keyframes}
		\label{MSCKF:line:descriptor}
		\For{feature tracks disappearing in the frame}
		\State Triangulate the point landmark
		\label{MSCKF:line:triangulate}
		\State Compute residuals and Jacobians of reprojection errors
		\State Cancel out Jacobians for the landmark
		\label{MSCKF:line:canceljac}
		
		\State Remove outliers with Mahalanobis test
		\label{MSCKF:line:mahal}
		\EndFor
		\State Form the whole residual vector and Jacobian
		\State Update state and covariance
		\label{MSCKF:line:update}
		\State \textit{Detect redundant frames}
		\label{MSCKF:line:detect-dud-frames}
		\State \textit{Update with reprojection errors in these frames}
		\label{MSCKF:line:update-with-dud-frames}
		\State \textit{Remove associated state variables and covariance entries}
		\label{MSCKF:line:remove-dud-variables}
		\EndWhile
	\end{algorithmic}
\end{algorithm}

\subsection{Consistency}
Despite easy interpretation and intuitive covariance tuning,
the chosen navigation error state is trajectory-dependent.
According to \cite{brossardExploitingSymmetriesDesign2018}, 
because the evolution of the chosen position error \eqref{eq:delta_t_WB} 
and velocity error \eqref{eq:delta_v_WB}
depend on the trajectory of the navigation state \cite[(A.27), (A.28)]{huai_collaborative_2017},
naively using the last estimate of the state vector to evaluate Jacobians for
\eqref{eq:imu_prop}, \eqref{eq:reprojection}, and \eqref{eq:reprojection_anchor}
will falsely eliminate the unobservable rotation about gravity
(\cite{li_high_2013, heschConsistencyAnalysisImprovement2014}) and lead to
optimistic covariance inconsistent to the actual estimator error.

To ensure consistency, Jacobians relative to navigation state variables
appearing in propagation \eqref{eq:process_jac} and measurements
\eqref{eq:reprojection} and \eqref{eq:reprojection_anchor} are evaluated
at propagated values of position and velocity, \ie, "first estimates"
\cite{li_high_2013},
and at last estimates of other variables.
Unlike \cite{geneva_openvins_2019} and \cite{engel_direct_2018},
we do not lock the evaluation points of rotation in computing Jacobians
because our chosen rotation error \eqref{eq:theta_WB} does not affect the
nullspace corresponding to the unobservable directions.

Jacobians of all other state variables are evaluated at the last estimate of
the state vector because they, \eg, IMU biases, time offsets, 
rolling shutter effect, do not affect the nullspace \cite{brossardExploitingSymmetriesDesign2018}.

\section{Simulation Study}
\label{sec:simulation}
We evaluate the KSF with simulation to
show that the designed method reports covariance consistent
to the actual estimation error and that the introduced sensor parameters are observable.
The simulation evaluates the proposed KSF in terms of consistency and
state estimation accuracy in comparison to a recent VIO method, OKVIS~\cite{leutenegger_keyframe_2015}.

\textbf{Error Metrics} The estimator consistency is measured by the Normalized Estimation
Error Squared (NEES) of components of the navigation state variable.
The expected value of NEES for a variable is its degrees of freedom, 
hence 3 for position error $\delta \mathbf{t}_{WB}$, 
3 for orientation error $\delta \boldsymbol{\theta}_{WB}$, 
and 6 for pose error $\delta\mathbf{T}_{WB} = (\delta \mathbf{t}_{WB},\delta \boldsymbol{\theta}_{WB})$.
An inconsistent estimator will optimistically estimate the covariance,
thus the computed NEES is much greater than its expected value \cite{li_high_2013}.
Following \cite[(3.7.6-1)]{bar-shalomEstimationApplicationsTracking2004}, with $n_s$ successful runs of an estimator,
the NEES $\boldsymbol{\epsilon}$ for position, orientation, and pose at epoch $t$ are 
\begin{align}
\boldsymbol{\epsilon}_{X}(t) = \frac{1}{n_s}\Sigma_{i=1}^{n_s} \delta
\mathbf{X}(t)^\intercal \boldsymbol{\Sigma}_{X}^{-1}(t) \delta \mathbf{X}(t)
\end{align}
where $\mathbf{X} = \mathbf{t}_{WB}, \boldsymbol{\theta}_{WB}, \mathbf{T}_{WB}$
and $\boldsymbol{\Sigma}_{X}$ is its covariance.

The accuracy of the estimated state is measured by Root Mean Square Error (RMSE) for components of the state vector.
For a component $\mathbf{X}$, the RMSE $r_X$ at epoch $t$ is computed by 
\begin{equation}
	r_X(t) = \sqrt{\frac{1}{n_s}\Sigma_{i=1}^{n_s} \delta \mathbf{X}(t)^{\intercal} \delta \mathbf{X}(t)}
\end{equation}

\begin{figure}[]
	\centering
	\subcaptionbox{}{\includegraphics[width=0.9\columnwidth]{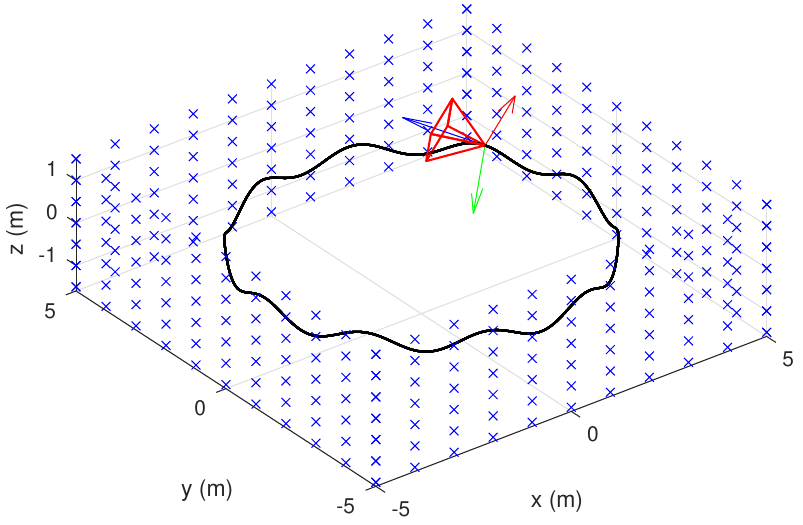}}%
	\\
	\subcaptionbox{}{\includegraphics[width=0.9\columnwidth]{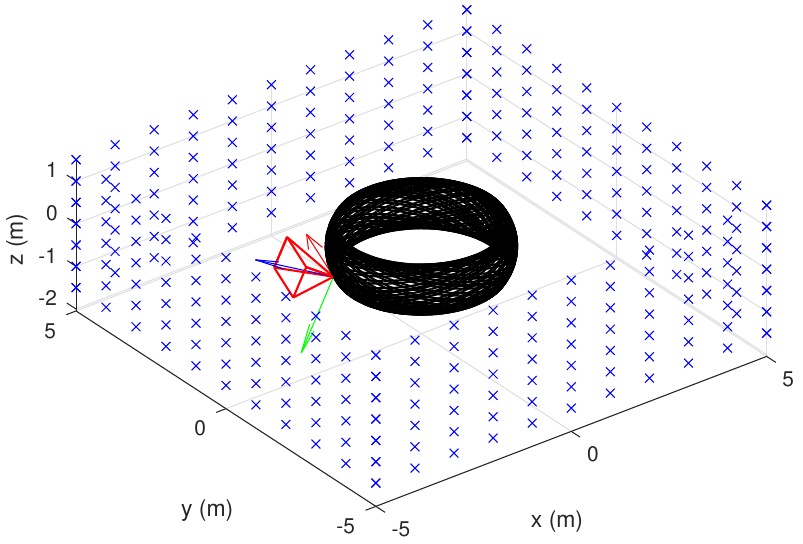}}
	\caption{The simulated scene with general wave motion (a) and torus motion (b) for five minutes.}
	\label{fig:scenarios}
\end{figure}

\textbf{Simulation Setup}
A scene with point landmarks distributed on four walls was simulated.
A camera-IMU platform traversed the scene for five minutes
with two motion types: a wavy circle (wave for short), and a yarn torus (Fig.~\ref{fig:scenarios}).
For the two motion types, the platform moved at an average velocity 1.26 m/s and 2.30 m/s, respectively.

\begin{table}[]
\centering
\caption{The covariance of the zero-mean Gaussian distributions from which
discrete noise samples are drawn. $f$ is the IMU sampling frequency.}
\begin{tabular}{lll}
	\hline
	$\Sigma$ & Gyroscope  & Accelerometer \\ \hline
	\begin{tabular}[c]{@{}l@{}}Bias White\\ Noise \\ 	\cite[(9) and (51)]{woodman2007introduction} \end{tabular} &
	\begin{tabular}[c]{@{}l@{}}$\sigma_{bg}^2 / f \mathbf{I}_3$ with\\
	$\sigma_{bg}=2\cdot10^{-5}$\\ $rad/s^2/\sqrt{Hz}$\end{tabular} &
	\begin{tabular}[c]{@{}l@{}}$\sigma_{ba}^2 / f \mathbf{I}_3$ with\\
	$\sigma_{ba} = 5.5\cdot10^{-5}$\\ $m/s^3/\sqrt{Hz}$\end{tabular} \\ \hline
	\begin{tabular}[c]{@{}l@{}}White Noise \\ \cite[(50)]{woodman2007introduction} \end{tabular}&
	\begin{tabular}[c]{@{}l@{}}$\sigma_g^2 f \mathbf{I}_3$ with\\
	$\sigma_g=1.2\cdot10^{-3}$\\ $rad/s/\sqrt{Hz}$\end{tabular}&
	\begin{tabular}[c]{@{}l@{}}$\sigma_a^2 f \mathbf{I}_3$ with\\ 
	$\sigma_a = 8\cdot10^{-3}$\\ $m/s^2/\sqrt{Hz}$\end{tabular} \\ \hline
\end{tabular}
\label{tab:imu_noise}
\end{table}

The simulated platform had a camera and an IMU. The camera captured images of size $752 \times 480$ at 10Hz.
The rolling shutter effect was simulated by solving for $t_{i,j}^k$  \eqref{eq:feature_time} as 
a root of the projection model \eqref{eq:reprojection} with the Newton-Raphson method.
For OKVIS, the rolling shutter effect was not added in simulation.
The image observations were corrupted by radial tangential distortion and
white Gaussian noise of 1 pixel standard deviation at each dimension.
The simulated inertial measurements were sampled at $f$=100 Hz,
corrupted by biases, white noise on biases and additive white noise.
Discrete noise samples were drawn from Gaussian distributions
tabulated in Table~\ref{tab:imu_noise}.
These noise parameters were chosen to be realistic for the IMU of 
a consumer smartphone.
The other parameters for generating the image and inertial data were set to
their mean values provided in Table~\ref{tab:init_value_std}.

\begin{table}
\centering
\caption{The IMU and camera parameters in KSF were initialized with samples
drawn from Gaussian distributions with mean values and standard
deviation $\sigma$ at each dimension.
$f_x$, $f_y$, $c_x$, $c_y$ are the camera projection parameters,
$k_1$, $k_2$, $p_1$, $p_2$ are the radial tangential distortion parameters, 
$t_d^0$ and $t_r^0$ are time offset and frame readout time of the camera.}
\begin{tabular}{cccccc}
\hline
 &
$\mathbf{b}_g (^{\circ}/s)$ &
\begin{tabular}[c]{@{}c@{}}$\mathbf{b}_a$\\ $(m/s^2)$\end{tabular} &
$\mathbf{T}_g (1)$ &
$\mathbf{T}_s (\frac{rad/s}{m/s^2})$ &
$\mathbf{T}_a (1)$ \\ \hline
Mean &
$\mathbf{0}$ &
$\mathbf{0}$ &
$\mathbf{I}_3$ &
$\mathbf{0}_{3\times3}$ &
$\mathbf{I}_3$ \\ \hline
\begin{tabular}[c]{@{}c@{}}$\sigma$\end{tabular} &
0.29 &
0.02 &
0.005 &
0.001 &
0.005 \\ \hline\hline
 &
$\mathbf{R}_{CB}$ &
\begin{tabular}[c]{@{}c@{}}$\mathbf{t}_{C0B}$\\ $(cm)$\end{tabular} &
\begin{tabular}[c]{@{}c@{}}$f_x, f_y$\\ $c_x, c_y$\\ $(pixel)$\end{tabular} &
\begin{tabular}[c]{@{}c@{}}$k_1, k_2$\\ $p_1, p_2$\end{tabular} &
\begin{tabular}[c]{@{}c@{}}$t_d^0 \enskip t_r^0$\\ $(ms)$\end{tabular} \\ \hline
Mean &
\begingroup
\setlength\arraycolsep{0.8pt}
\renewcommand*{\arraystretch}{0.4}
$\begin{bmatrix}0 & -1 & 0 \\ 0 & 0 & -1 \\ 1 & 0 & 0\end{bmatrix}$
\endgroup &
$\mathbf{0}$ &
\begin{tabular}[c]{@{}c@{}}350, 360 \\ 378, 238\end{tabular} &
\begin{tabular}[c]{@{}c@{}}0, 0 \\ 0, 0\end{tabular} &
\begin{tabular}[c]{@{}c@{}}500\\ 20\end{tabular} \\ \hline
\begin{tabular}[c]{@{}c@{}}$\sigma$\end{tabular} &
0 &
2.0 &
\begin{tabular}[c]{@{}c@{}}5.0, 5.0\\ 5.0, 5.0\end{tabular} &
\begin{tabular}[c]{@{}c@{}}0.05,0.01\\ 0.001,0.001\end{tabular} &
\begin{tabular}[c]{@{}c@{}}5\\ 5\end{tabular} \\ \hline
\end{tabular}
\label{tab:init_value_std}
\end{table}

\textbf{Estimator Setup} A simulation frontend was created to provide feature tracks to an
estimator, \ie, the backend of KSF or OKVIS.
The frontend associated observations of a landmark between consecutive
frames and between current frame and the last keyframe.
For torus / wave motion, the average feature track length was 5.8 / 6.8, 
and the average number of observed landmarks in an image was 40.5 / 60.5.

An estimator were initialized with the true pose but a noisy velocity estimate
corrupted with noise sampled from Gaussian distribution 
$N(\mathbf{0}, 0.05^2 \mathbf{I}_3 \hspace{0.2em} m^2/s^4)$.
For KSF, the initial values for camera and IMU parameters were drawn
from Gaussian distributions given in Table~\ref{tab:init_value_std}.
For OKVIS, the initial biases were drawn from distributions given in
Table~\ref{tab:init_value_std} and the other parameters were locked to
mean values in Table~\ref{tab:init_value_std} except for time delay
and readout time of the camera which were set zero.

Each estimator ran 1000 times for each motion type. All runs were finished
successfully, \ie, with the error in position $\le$ 100 m at the end.

\begin{figure}[]
	\centering
	\includegraphics[width=0.9\columnwidth]{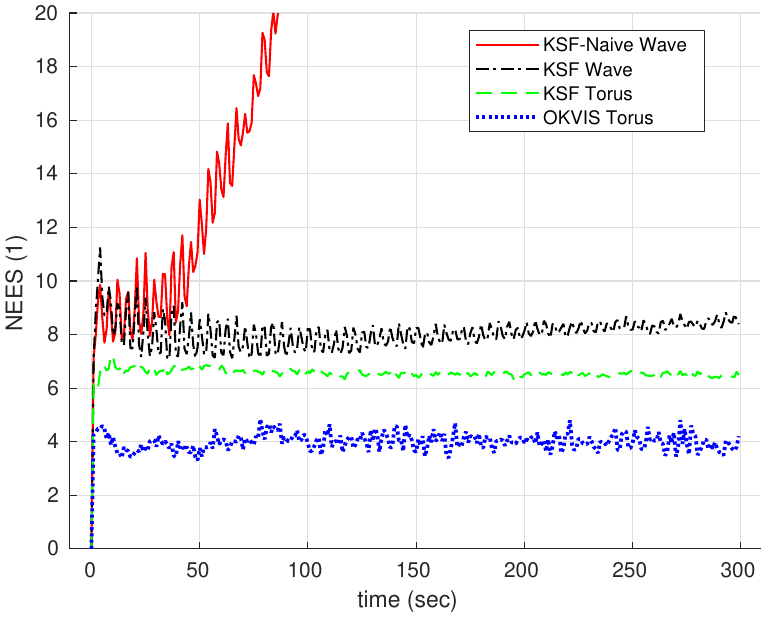}%
	\caption{The history of NEES for poses estimated by 
		KSF with naive Jacobians under wave motion (KSF-Naive Wave), 
	KSF with First Estimate Jacobians (FEJ) under wave motion (KSF Wave), KSF with FEJ under torus motion (KSF Torus), and OKVIS under torus motion (OKVIS Torus).
	Expected values of NEES for position, orientation, and pose are 3, 3, and 6.}
	\label{fig:nees}
\end{figure}

\textbf{Estimator Consistency}
We compared the NEES for KSF through wave motion with Jacobians naively
evaluated at last estimates, KSF through wave motion with first estimate
Jacobians (FEJ) by default, and KSF through torus motion with FEJ by default.
The NEES for OKVIS was not computed because of rank deficient Jacobians in the
estimator.
The evolution of NEES is visualized in Fig.~\ref{fig:nees}. 
The NEES values averaged over the last 10 seconds to smooth out jitters are
tabulated in Table~\ref{tab:nees_rmse}.

We had observed that the NEES could be reduced closer to its expected value by
decreasing the variance of noise in sampling IMU readings
while keeping the noise variance unchanged in the VIO estimator.
Doing so is somewhat reasonable in the sense that the simulation setup has unmodeled
quantization errors.
However, for simplicity, the following results were obtained without weakening noise added to IMU readings.

From curves in Fig.~\ref{fig:nees} and final values of NEES in Table~\ref{tab:nees_rmse},
we see that the proposed estimator, KSF with FEJ, is consistent in terms of NEES.
Surprisingly, OKVIS also achieves consistent NEES though nonlinear factors are always linearized at the latest variable estimates.
The results also indicate that NEES is affected to some extent by motion type
as the motion that excited the IMU in diverse ways led NEESs closer to
expected values.

\begin{table}[]
\caption{NEES and RMSE computed over 1000 runs for OKVIS with wave motion and no systematic error,
KSF with naive Jacobians and wave motion,
KSF with first estimate Jacobians (FEJ) and wave motion, and KSF with FEJ and
torus motion. RMSE values are 0 at the start because poses were
initialized with ground truth.}
\begin{tabular}{l|lll|ll|}
\cline{2-6}
& \multicolumn{3}{l|}{\begin{tabular}[c]{@{}l@{}}NEES averaged over\\
last 10 seconds\end{tabular}} & \multicolumn{2}{l|}{RMSE at end} \\ \cline{2-6} 
& \begin{tabular}[c]{@{}l@{}}Position\\(1)\end{tabular} & \begin{tabular}[c]{@{}l@{}}Orienta-\\tion (1)\end{tabular}     
& \begin{tabular}[c]{@{}l@{}}Pose\\(1)\end{tabular}& \begin{tabular}[c]{@{}l@{}}Position\\ (m)\end{tabular} &
\begin{tabular}[c]{@{}l@{}}Orienta- \\ tion ($^{\circ}$)\end{tabular} \\ \hline
\multicolumn{1}{|l|}{\begin{tabular}[c]{@{}l@{}}Theoretic\\ value\end{tabular}}
& 3      & 3 & 6    & N/A    & N/A \\ \hline
\multicolumn{1}{|l|}{\begin{tabular}[c]{@{}l@{}}KSF-Naive\\Wave\end{tabular}}     & 11.28  & 146.94    & 150.80       & 0.72  & 8.48       \\ \hline
\multicolumn{1}{|l|}{KSF Wave}       &4.53 &    5.13  &  8.50 & 0.55  & 5.52     \\ \hline
\multicolumn{1}{|l|}{KSF Torus}       &3.32 &    3.11 &    6.46    & 0.43   & 4.46       \\ \hline
\multicolumn{1}{|l|}{OKVIS Torus}   & 0.85   &   2.95  &    3.80  & 1.03 & 5.13 \\ \hline
\end{tabular}
\label{tab:nees_rmse}
\end{table}

\textbf{Estimation of Pose and Sensor Parameters}
From the final RMSE in Table~\ref{tab:nees_rmse}, we see that KSF outperformed
OKVIS in pose accuracy, and the inconsistent KSF with naively computed Jacobians
led to worse pose estimates. 

For KSF with wave and torus motion, the RMSEs of sensor parameters at different
epochs are tabulated in Table~\ref{tab:sim_state_estimate}.
For both motion types, all sensor parameter components 
decreased in uncertainty over time as observed in
\cite{li_calibration_2014, huai_collaborative_2017}, though at varying rates.
The frame readout time $t_r$ converged in general but became stagnant
after its estimate was within a radius of the true value.

The uncertainty of IMU parameters decreased faster with torus
motion than with the less diverse wave motion.
Obviously, for wave motion (2nd row block of Table~\ref{tab:sim_state_estimate}),
accelerometer bias $\mathbf{b}_a$ and gyroscope bias $\mathbf{b}_g$ converged very slowly.
Besides the simple motion, we suspected that inclusion of IMU
parameters, \ie, $\mathbf{T}_g$, $\mathbf{T}_s$, and $\mathbf{T}_a$, in the
estimator also slowed the convergence of IMU biases.
To verify this, we ran KSF through wave motion while locking additional IMU
parameters to their nominal values.
The resultant RMSEs are given in the first row block of Table~\ref{tab:sim_state_estimate}.
From the results, we see that when the IMU systematic errors,
$\mathbf{T}_g$, $\mathbf{T}_s$, and $\mathbf{T}_a$, were removed, estimates for
biases converged much faster, and many camera parameters also sped up in
convergence.

\begin{table*}[]
\centering
\caption{RMSE of sensor parameters at several epochs computed from 1000
simulation runs of KSF with first estimate Jacobians by default in three settings, 
KSF through wave motion with IMU parameters except biases locked at nominal values, 
KSF through wave motion, KSF through torus motion.}
\label{tab:sim_state_estimate}
\begin{tabular}{l|ccccccccccccc}
	\hline
	\begin{tabular}[c]{@{}l@{}}Test \\ setup\end{tabular} & \multicolumn{1}{l}{\begin{tabular}[c]{@{}l@{}}Time\\ (s)\end{tabular}} &
	\multicolumn{1}{l}{\begin{tabular}[c]{@{}l@{}}$\mathbf{b}_g$\\ $(^{\circ}/s)$\end{tabular}} &
	\multicolumn{1}{l}{\begin{tabular}[c]{@{}l@{}}$\mathbf{b}_a$\\ $(m/s^2)$\end{tabular}} &
	\multicolumn{1}{l}{\begin{tabular}[c]{@{}l@{}}$\mathbf{T}_g$\\ $(0.001)$\end{tabular}} &
	\multicolumn{1}{l}{\begin{tabular}[c]{@{}l@{}}$\mathbf{T}_s (0.001$\\ $\frac{rad/s}{m/s^2})$\end{tabular}} &
	\multicolumn{1}{l}{\begin{tabular}[c]{@{}l@{}}$\mathbf{T}_a$\\ $(0.001)$\end{tabular}} &
	\multicolumn{1}{l}{\begin{tabular}[c]{@{}l@{}}$\mathbf{t}_{C0B}$ \\ $(cm)$\end{tabular}} &
	\multicolumn{1}{l}{\begin{tabular}[c]{@{}l@{}}$(f_x, f_y)$\\ $(pixel)$\end{tabular}} &
	\multicolumn{1}{l}{\begin{tabular}[c]{@{}l@{}}$(c_x, c_y)$\\ $(pixel)$\end{tabular}} &
	\multicolumn{1}{l}{\begin{tabular}[c]{@{}l@{}}$(k_1, k_2)$\\ $(0.001)$\end{tabular}} &
	\multicolumn{1}{l}{\begin{tabular}[c]{@{}l@{}}$(p_1, p_2)$\\ $(0.001)$\end{tabular}} &
	\multicolumn{1}{l}{\begin{tabular}[c]{@{}l@{}}$t_d^0$\\ $(ms)$\end{tabular}} &
	\multicolumn{1}{l}{\begin{tabular}[c]{@{}l@{}}$t_r^0$\\ $(ms)$\end{tabular}} \\ \hline
	\multirow{6}{*}{\begin{tabular}[c]{@{}l@{}}Wave \\ with\\ fixed\\
	$\mathbf{T}_g$\\ $\mathbf{T}_s$\\ $\mathbf{T}_a$\end{tabular}}
    & 0.0 & 0.47 & 0.034 & 0.00 & 0.00 & 0.00 & 3.45 & 7.36 & 7.07 & 55.89 & 1.44 & 4.34 & 4.93 \\ \cline{2-14} 
	& 3.0 & 0.21 & 0.029 & 0.00 & 0.00 & 0.00 & 1.79 & 1.33 & 1.28 & 3.84 & 0.96 & 0.54 & 0.81 \\ \cline{2-14} 
	& 10.0 & 0.08 & 0.025 & 0.00 & 0.00 & 0.00 & 0.94 & 0.58 & 0.65 & 2.17 & 0.56 & 0.27 & 0.54 \\ \cline{2-14} 
	& 30.0 & 0.05 & 0.016 & 0.00 & 0.00 & 0.00 & 0.61 & 0.32 & 0.43 & 1.14 & 0.32 & 0.16 & 0.42 \\ \cline{2-14} 
	& 100.0 & 0.03 & 0.008 & 0.00 & 0.00 & 0.00 & 0.32 & 0.17 & 0.21 & 0.60 & 0.17 & 0.09 & 0.39 \\ \cline{2-14} 
	& 300.0 & 0.02 & 0.005 & 0.00 & 0.00 & 0.00 & 0.20 & 0.09 & 0.13 & 0.36 & 0.10 & 0.06 & 0.39 \\ \hline\hline
	\multirow{6}{*}{Wave} & 0.0 & 0.50 & 0.035 & 14.92 & 3.04 & 14.83 & 3.45 & 7.26 & 7.02 & 49.82 & 1.40 & 4.96 & 4.89 \\ \cline{2-14} 
	& 3.0 & 0.43 & 0.033 & 10.59 & 1.89 & 12.77 & 2.16 & 1.90 & 1.70 & 4.24 & 1.12 & 1.08 & 0.93 \\ \cline{2-14} 
	& 10.0 & 0.42 & 0.032 & 6.81 & 1.13 & 9.67 & 1.48 & 1.04 & 1.47 & 2.17 & 1.00 & 0.45 & 0.56 \\ \cline{2-14} 
	& 30.0 & 0.40 & 0.032 & 5.11 & 0.92 & 8.09 & 1.20 & 0.63 & 1.35 & 1.23 & 0.89 & 0.27 & 0.43 \\ \cline{2-14} 
	& 100.0 & 0.35 & 0.031 & 3.44 & 0.74 & 6.28 & 0.83 & 0.40 & 0.96 & 0.67 & 0.64 & 0.15 & 0.40 \\ \cline{2-14} 
	& 300.0 & 0.29 & 0.031 & 2.21 & 0.60 & 5.10 & 0.56 & 0.26 & 0.58 & 0.40 & 0.39 & 0.09 & 0.39 \\ \hline\hline
	\multirow{6}{*}{Torus} & 0.0 & 0.50 & 0.034 & 14.90 & 3.01 & 14.91 & 3.48 & 7.16 & 6.94 & 51.38 & 1.48 & 5.13 & 5.02 \\ \cline{2-14} 
	& 3.0 & 0.44 & 0.032 & 13.06 & 2.45 & 13.03 & 2.97 & 2.82 & 2.20 & 3.69 & 1.22 & 3.21 & 2.28 \\ \cline{2-14} 
	& 10.0 & 0.39 & 0.032 & 6.92 & 1.37 & 9.46 & 2.07 & 1.44 & 1.44 & 1.75 & 0.91 & 1.11 & 1.11 \\ \cline{2-14} 
	& 30.0 & 0.22 & 0.028 & 4.18 & 0.70 & 7.02 & 1.52 & 0.72 & 1.17 & 0.95 & 0.74 & 0.49 & 0.69 \\ \cline{2-14} 
	& 100.0 & 0.12 & 0.023 & 2.45 & 0.38 & 4.80 & 0.93 & 0.39 & 0.79 & 0.54 & 0.51 & 0.27 & 0.52 \\ \cline{2-14} 
	& 300.0 & 0.08 & 0.017 & 1.56 & 0.23 & 3.25 & 0.60 & 0.23 & 0.53 & 0.31 & 0.37 & 0.16 & 0.51 \\ \hline
\end{tabular}
\end{table*}

\section{Real-World Experiments}
\label{sec:real-world-tests}
With real-world data, the proposed KSF was evaluated in
four groups, sensitivity of pose accuracy to several design options,
$N$-camera-IMU platform calibration in comparison to reference values,
accuracy of motion estimation relative to recent VIO methods, 
and computation cost.

The pose accuracy was measured with Root Mean Square (RMS) of the Absolute Translation Error (ATE)
\cite[(24)]{zhang2018tutorial}, the Relative Rotation Error (RRE)
\cite[(27)]{zhang2018tutorial}, and the Relative Translation Error (RTE)
\cite[(27)]{zhang2018tutorial} as implemented by \cite{zhang2018tutorial}.
ATE in units of meters measures the overall deviation of the estimated
pose sequence from ground truth in translation. RRE in units $^\circ/m$ and RTE
in units \% quantifies deviations from ground truth over
trajectory segments of fixed lengths, in rotation and translation,
respectively.
To mitigate the random behavior of a VIO estimator, results from three runs were
used to compute the error metrics since empirically metrics computed from three
runs changed little, \eg, less than 0.05\% in ATE.
To compute relative errors, RRE and RTE, distance intervals 
[40, 60, 80, 100, 120] were used as in the UZH-FPV dataset~\cite{delmerico2019we}.

RRE and RRE are invariant to a rigid transformation of the pose sequence.
However, ATE is dependent on the spatial alignment. As recommended in
\cite[Table~I]{zhang2018tutorial}, we aligned the trajectory estimated by a VIO
method to the
ground truth with a transformation of a yaw-only rotation and a 3D translation.
This 4-DOF alignment requires that poses in the estimated trajectory and
ground truth are expressed in world frames with z-axis aligned to gravity. 
The first part is fulfilled since all VIO methods used for
comparison estimate poses of the IMU sensor frame in a
gravity-aligned world frame. 
The second part is fulfilled by benchmarks, EuRoC~\cite{burri2016euroc} and TUM VI~\cite{schubert2018tum}.

For the following tests, we initialized a VIO estimator with 
the standard deviation of position, orientation error, velocity, 
gyroscope bias, and accelerometer bias,
$0.01 \mathbf{I}_3 \hspace{0.2em}  m$,
$\mathrm{diag}(1^{\circ}, 1^{\circ}, 3^{\circ})$,
$0.1 \mathbf{I}_3 \hspace{0.2em} m/s$,
$1.72 \mathbf{I}_3 \hspace{0.2em} {}^{\circ}/s$, 
and $0.1 \mathbf{I}_3 \hspace{0.2em} m/s^2$, respectively.

\subsection{Sensitivity Study}
\label{subsec:sensitivity}
By tuning KSF on the EuRoC dataset, we studied its dependence on the feature tracking method,
its sensitivity to key parameters including number of keyframes
$N_{kf}$ and number of temporally successive frames $N_{tf}$,
and the effect on pose estimation of landmark parameterization and 
sensor calibration.
The entire EuRoC dataset~\cite{burri2016euroc} without modification was used
because it is less demanding than other benchmarks, \eg, TUM VI~\cite{schubert2018tum},
and we feel that a challenging dataset may result in parameters not well
suitable to the vast average data.

We tested the parameter sensitivity of KSF in the monocular setting
because most findings are expected to carry over to $N$-camera settings.
By default, KSF processed data synchronously in the non-realtime fashion.

For each subsequent sensitivity test, we began with a
manually selected baseline combination of parameters which was expected
to achieve good pose estimation accuracy based on our experiences.
Then we ran KSF with components or parameters of interest differing from the baseline.
In the end, the optimal component or parameter combination was identified 
by looking at the pose estimation accuracy of different combinations.

\textbf{Feature Tracking Method}
To show the merit of keyframe-based feature matching, we assessed 
KSF with alternative feature association methods including 
frame-wise KLT feature tracking \cite{bradski2008learning} (hence KSF-KLT), 
and sequential frame matching by BRISK descriptors (hence KSF-seq).
The relative errors are summarized in \ref{tab:RE_feature_matching},
confirming the advantage of keyframe-based feature association.
Visual check of the superimposed trajectory plots
indicated that both KSF-KLT and KSF-seq often had jumps in trajectory when 
the Micro Aerial Vehicle (MAV) was nearly stationary.
This was expected because these two alternatives got mainly uninformative feature
observations of small disparities during standstill.

\begin{table}[]
\centering
\caption{Relative errors of KSF on EuRoC dataset with keyframe-based, KLT, and
sequential feature
association and of KSF with varying number of keyframes
(first number after dash) and temporal frames (last number after dash).}
\label{tab:RE_feature_matching}
\begin{tabular}{lll}
\hline & RTE (\%) &  RRE ($^{\circ}/m$) \\ \hline
KSF &     0.73 &  0.030 \\ \hline
KSF-KLT &     3.76 &  0.053 \\ \hline
KSF-seq &     0.87 &  0.034 \\ \hline\hline
KSF-3-3 &     3.42 &  0.047 \\ \hline
KSF-5-5 &     0.81 &  0.032 \\ \hline
KSF-7-5 &     0.72 &  0.030 \\ \hline
KSF-7-7 &     0.73 &  0.031 \\ \hline
KSF-9-5 &     0.71 &  0.030 \\ \hline
KSF-9-9 &     0.69 &  0.031 \\ \hline
\end{tabular}
\end{table}

\textbf{Number of Keyframes and Temporal Frames}
The backend sliding window consists of $N_{kf}$ keyframes and $N_{tf}$ recent temporal frames.
A larger sliding window is expected to give better pose estimates 
but incurs higher computation as shown in \ref{tab:RE_feature_matching}.
To balance accuracy and resource consumption, we selected 7 keyframes and 5
temporal frames for KSF in subsequent tests.

\textbf{Lock Sensor Parameters}
As the proposed method involves quite a few sensor parameters, 
we evaluated the effect of calibrating sensor parameters on pose accuracy.
Starting from an extreme baseline, unlocking (or locking) all sensor parameters,
we ran KSF with only one component of the sensor parameters locked (or unlocked).
Remarkably, the IMU biases were always estimated.
Then the relative errors for results of different calibration settings were
computed (Table~\ref{tab:sensitivity-calibration}).
From the first half of Table~\ref{tab:sensitivity-calibration}, we see that
locking IMU parameters 
$\{\mathbf{T}_g, \mathbf{T}_s, \mathbf{T}_a\}$ or camera intrinsic parameters
can improve pose accuracy while locking camera extrinsic parameters have a
slight impact, and locking camera temporal parameters worsens the accuracy.
From the second half of Table~\ref{tab:sensitivity-calibration}, we see that
estimating IMU or camera intrinsic parameters adversely affects the results, 
but estimating camera extrinsic or temporal parameters improves the accuracy.
Interestingly, Yang \etal \cite[Table V]{yangOnlineImuIntrinsic2020} 
reported that online IMU calibration marginally worsens RMS ATE values
for many EuRoC sequences and they reasoned that the MAV motion is not diverse enough.
In summary, for the EuRoC dataset, locking IMU and camera intrinsic parameters
while calibrating camera extrinsic and temporal parameters led to better pose accuracy.
This finding is likely due to good individual calibration of the IMU and cameras used for
capturing the EuRoC dataset.

\begin{table}
\centering
\caption{Relative errors of KSF on EuRoC dataset when estimating subsets of sensor parameters,
including IMU parameters $\{\mathbf{T}_g, \mathbf{T}_s, \mathbf{T}_a\}$, camera
extrinsic ($\mathbf{t}_{C0B}$), intrinsic (projection and distortion), and temporal
(time delay $t_d^0$ and frame readout time) parameters.}
\label{tab:sensitivity-calibration}
\begin{tabular}{lll}
\hline & RTE (\%) &  RRE ($^{\circ}/m$) \\ \hline
KSF unlock all &     0.79 &  0.037  \\ \hline
KSF lock $\mathbf{T}_g\, \mathbf{T}_s \, \mathbf{T}_a$ 
&     0.73 &  0.031 \\ \hline
KSF lock extrinsic &     0.78 &  0.036 \\ \hline
KSF lock intrinsic &     0.72 &  0.031 \\ \hline
KSF lock temporal &     0.87 &  0.036 \\ \hline \hline
KSF lock all &     0.71 &  0.029 \\ \hline
KSF unlock $\mathbf{T}_g\, \mathbf{T}_s \, \mathbf{T}_a$ &
0.75 &  0.032 \\ \hline
KSF unlock extrinsic &     0.69 &  0.030 \\ \hline
KSF unlock intrinsic &     0.73 &  0.029 \\ \hline
KSF unlock $t_d^0$ &     0.70 &  0.029 \\ \hline	
\end{tabular}
\end{table}

\textbf{Landmark Parameterization}
It was observed in \cite{li_high_2013, sola2012impact} that anchored inverse
depth (AID) parameterization \eqref{eq:reprojection_anchor} for landmarks
outperformed the homogeneous point (HP) parameterization \eqref{eq:reprojection}.
This was confirmed in our tests by relative errors for KSF with AID by default
versus with HP in Table~\ref{tab:sensitivity-miscell}.

\textbf{$N$-Camera Setup and Real-Time Processing}
We also evaluated KSF with input from a stereo camera (KSF-n), KSF running in
asynchronous mode (KSF-async), \ie, real-time processing, and KSF-n in
asynchronous mode (KSF-n-async).
The relative errors are given in Table~\ref{tab:sensitivity-miscell} where
we see the benefit of stereo cameras in reducing the relative translation error.
Surprisingly, asynchronous processing led to slightly better result
than synchronous processing.
But this trend was not observed on other benchmarks, \eg, TUM VI.
Aside from the random behavior of the algorithm, another cause could be that
some of the rich feature descriptors in the EuRoC dataset were redundant and
discarding them might be better regarding pose accuracy 
(see also \cite{engel_direct_2018}).

\begin{table}
	\centering
\caption{Pose accuracy of KSF on EuRoC dataset with an alternative landmark
parameterization, Homogeneous Point (HP) in the world frame,
with asynchronous real-time processing (async), with asynchronous processing of stereo images
(n-async), and with non-realtime processing of stereo images (n). 
By default, KSF processes images from a single camera synchronously
with anchored inverse depth parameterization.}
	\label{tab:sensitivity-miscell}
	\begin{tabular}{lll}
		\hline & RTE (\%) &  RRE ($^{\circ}/m$) \\ \hline
		KSF &     0.68 &  0.031 \\ \hline
		KSF-HP &     0.71 &  0.029 \\ \hline
		KSF-async &     0.67 &  0.030 \\ \hline
		KSF-n-async &     0.39 &  0.028 \\ \hline
		KSF-n &     0.40 &  0.029 \\ \hline
	\end{tabular}
\end{table}

\subsection{Sensor Calibration}
\label{subsec:sensor-calibration}
To show the capability of KSF in estimating the extrinsic, intrinsic, and
time delay parameters of a $N$-camera-IMU platform, KSF was used to process several raw
sequences of TUM VI~\cite{schubert2018tum} and the estimated sensor parameters were compared to
provided reference values.
Moreover, to show the possibility of calibrating frame readout time,
sequences captured by a few Android phones were processed with KSF and the
estimated readout times were compared to values obtained with a LED panel.

\textbf{Stereo-Camera-IMU Calibration}
The TUM VI benchmark provided the raw sequences collected by the stereo-camera-IMU platform
which was subject to a variety of sensor systematic errors. On these raw
sequences with diverse motion, KSF was expected to give good estimates of these
systematic errors. On the other hand, the systematic errors of the platform had been
estimated by the authors with dedicated calibration sequences and reference
values of relevant parameters were provided.
Thus, by comparing the reference values and the estimates from KSF,
we can assess KSF's capability of sensor calibration.

To this end, we chose six room sessions of the raw TUM VI sequences (each about two-minute long)
because they contained rich motion and ground truth covering the full trajectory. Images in these raw data were
down sampled by \texttt{pyrDown} \cite{bradski2008learning} to 512$\times$512.
Both images and IMU data were timestamped with the \texttt{header.stamp}
field of raw messages. Otherwise these raw data were intact.
This contrasted with \cite{xiao2019online} where the initial IMU biases in
the raw data were corrected before online calibration.

KSF was initialized with nominal values and standard deviations for sensor
parameters as listed in Column 2, Table~\ref{tab:tumvi-room-calib}. 
Because the sensor pixel is usually square, the same focal length $f^k$
was used for both $x$ and $y$ directions of camera $k$.
The camera intrinsic parameters were initialized assuming we know their rough values.
For instance, $f^k$ was initialized with a rounded number 190.
The principal points were simply initialized at the image centers, \ie, (256, 256).
The IMU noise parameters were set as the stereo KSF for
processing the calibrated TUM VI sequences in Section
\ref{subsec:pose-estimation}.
Each sequence was processed three times to mitigate the nondeterministic behavior of
the algorithm. 
For six raw sequences three runs each, we computed means and standard deviations of
estimated sensor parameters at the end of these 18 trials.
The estimated parameters including IMU biases and systematic errors, and camera
extrinsic, intrinsic, and temporal parameters are given in
Table~\ref{tab:tumvi-room-calib}.

The calibration parameters coming with the TUM VI dataset were obtained by a global
bundle adjustment with special data captured while observing a calibration target.
These parameters followed the IMU and camera extrinsic models defined by TUM VI. 
For comparison, they needed to be converted to parameters of the IMU and camera extrinsic models used by KSF.
The conversion was derived by comparing TUM VI models with ours.
Given TUM VI IMU model coefficient matrices $\mathbf{M}_g$, $\mathbf{M}_a$,
IMU biases, $\mathbf{b}_{g, t}$ and $\mathbf{b}_{a,t}$, extrinsic parameters
of the two cameras in the TUM VI sensor frame $\{St\}$, $[\mathbf{R}_{StC0}, \mathbf{t}_{StC0}]$ and $[\mathbf{R}_{StC1},
\mathbf{t}_{StC1}]$, the parameters of KSF can be obtained by
\begin{align}
\mathbf{b}_g &= \mathbf{M}_g^{-1} \mathbf{b}_{g, t} \\
\mathbf{b}_a &= \mathbf{M}_a^{-1} \mathbf{b}_{a, t} \\
\mathbf{T}_g &= \mathbf{M}_g^{-1} \mathbf{R}_{StC0} \mathbf{R}_{SnC}^\intercal \\
\mathbf{T}_s &= \mathbf{0} \\
\mathbf{T}_a &= \mathbf{M}_a^{-1} \mathbf{R}_{StC0} \mathbf{R}_{SnC}^\intercal \\
\mathbf{p}_{C0B} &= - \mathbf{R}_{StC0}^\intercal \mathbf{t}_{StC0} \\
\mathbf{p}_{BC1} &= \mathbf{R}_{SnC} \mathbf{R}_{StC0}^\intercal \mathbf{t}_{StC1} \\
\mathbf{R}_{BC1} &= \mathbf{R}_{SnC} \mathbf{R}_{StC0}^\intercal \mathbf{R}_{StC1}
\end{align}
where $\mathbf{R}_{SnC}$ is the orientation between the camera frame and the nominal IMU frame $\{Sn\}$,
\begin{equation}
	\mathbf{R}_{SnC} = \begin{bmatrix}
	-1 & 0 & 0 \\
	0 & 0 & -1 \\
	0 & -1 & 0
	\end{bmatrix}.
\end{equation}
$\mathbf{R}_{SnC}$ is needed because KSF fixed orientation of the
body frame $\{B\}$ relative to the camera frame $\{C\}$ to
$\mathbf{R}_{SnC}$ in processing raw sequences of TUM VI,
\ie, starting with a coarse calibration.

\begin{table}[]
	\centering
	\caption{Mean and standard deviation of 18 estimated state vectors by running KSF in two configurations (calibrating all parameters except for frame readout time, calibrating all except for camera intrinsics and frame readout time), on six room sessions of TUM VI for three times. Each estimated state vector was the last one from a run. Note nominal values, 190 and (256, 256) had been subtracted from focal length and principal point. $\delta\boldsymbol{\theta}_{BC1}$ relates the nominal orientation and the estimated one such that $\mathbf{R}_{SnC} = \exp(\delta \boldsymbol{\theta}_{BC1}^\times) \hat{\mathbf{R}}_{BC1}$. $\mathbf{1}:=[1, 1, 1]^\intercal$.}
	\label{tab:tumvi-room-calib}
	\begin{tabular}{lllll}
		\hline
		\multirow{2}{*}{Parameter} &
		\multirow{2}{*}{\begin{tabular}[c]{@{}l@{}}Initial\\ Value\end{tabular}} &
		\multicolumn{2}{l}{Final Value} &
		\multirow{2}{*}{\begin{tabular}[c]{@{}l@{}}TUM VI\\ Ref.\end{tabular}} \\ \cline{3-4}
		&
		&
		calibrate all &
		\begin{tabular}[c]{@{}l@{}}fix camera\\ intrinsics\end{tabular} &
		\\ \hline
		$\mathbf{b}_g {}(^{\circ}/s)$ &
		\begin{tabular}[c]{@{}l@{}}$\mathbf{0} \pm$\\ $1.72\mathbf{1}$\end{tabular} &
		\begin{tabular}[c]{@{}l@{}}2.08 $\pm$ 0.31\\ 0.47 $\pm$ 0.13\\ 1.19 $\pm$ 0.36\end{tabular} &
		\begin{tabular}[c]{@{}l@{}}1.97 $\pm$ 0.21\\ 0.47 $\pm$ 0.14\\ 1.08 $\pm$ 0.30\end{tabular} &
		\begin{tabular}[c]{@{}l@{}}1.72\\ 0.38\\ 0.93\end{tabular} \\ \hline
		$\mathbf{b}_a (m/s^2)$ &
		$\mathbf{0} \pm 0.1\mathbf{1}$ &
		\begin{tabular}[c]{@{}l@{}}-1.26 $\pm$ 0.04\\ -0.30 $\pm$ 0.05\\ 0.44 $\pm$ 0.05\end{tabular} &
		\begin{tabular}[c]{@{}l@{}}-1.28 $\pm$ 0.04\\ -0.31 $\pm$ 0.04\\ 0.41 $\pm$ 0.03\end{tabular} &
		\begin{tabular}[c]{@{}l@{}}-1.30\\ -0.39\\ 0.38\end{tabular} \\ \hline
		\begin{tabular}[c]{@{}l@{}}$\mathbf{T}_g - \mathbf{I}$\\ $(0.001)$\end{tabular} &
		\begin{tabular}[c]{@{}l@{}}$\mathbf{0} \pm$\\ $5 \mathbf{1}_{3\times3}$\end{tabular} &
		\begin{tabular}[c]{@{}l@{}}58.27 $\pm$ 1.91\\ 27.90 $\pm$ 3.67\\ -7.44 $\pm$ 1.10\\ -24.78 $\pm$ 3.25\\ -87.05 $\pm$ 0.86\\ 30.46 $\pm$ 2.33\\ 7.87 $\pm$ 0.99\\ -36.82 $\pm$ 3.31\\ -18.05 $\pm$ 1.34\end{tabular} &
		\begin{tabular}[c]{@{}l@{}}58.47 $\pm$ 1.26\\ 30.53 $\pm$ 1.47\\ -8.33 $\pm$ 0.58\\ -26.95 $\pm$ 1.15\\ -87.69 $\pm$ 0.69\\ 35.48 $\pm$ 2.67\\ 9.51 $\pm$ 1.05\\ -43.92 $\pm$ 3.29\\ -18.45 $\pm$ 0.83\end{tabular} &
		\begin{tabular}[c]{@{}l@{}}59.30\\ 30.09\\ -8.80\\ -27.24\\ -87.09\\ 34.08\\ 10.38\\ -41.27\\ -16.54\end{tabular} \\ \hline
		\begin{tabular}[c]{@{}l@{}}$\mathbf{T}_s$\\ $(0.001$\\ $\frac{rad/s}{m/s^2})$\end{tabular} &
		\begin{tabular}[c]{@{}l@{}}$\mathbf{0} \pm$\\ $\mathbf{1}_{3\times3}$\end{tabular} &
		\begin{tabular}[c]{@{}l@{}}0.49 $\pm$ 0.38\\ 0.05 $\pm$ 0.10\\ -0.67 $\pm$ 0.56\\ -0.05 $\pm$ 0.14\\ 0.19 $\pm$ 0.16\\ -0.11 $\pm$ 0.25\\ 0.43 $\pm$ 0.33\\ 0.10 $\pm$ 0.22\\ -0.44 $\pm$ 0.71\end{tabular} &
		\begin{tabular}[c]{@{}l@{}}0.29 $\pm$ 0.22\\ 0.07 $\pm$ 0.14\\ -0.47 $\pm$ 0.39\\ -0.11 $\pm$ 0.13\\ 0.05 $\pm$ 0.12\\ -0.12 $\pm$ 0.26\\ 0.12 $\pm$ 0.26\\ 0.03 $\pm$ 0.20\\ -0.26 $\pm$ 0.57\end{tabular} &
		\begin{tabular}[c]{@{}l@{}}0.00\\ 0.00\\ 0.00\\ 0.00\\ 0.00\\ 0.00\\ 0.00\\ 0.00\\ 0.00\end{tabular} \\ \hline
		\begin{tabular}[c]{@{}l@{}}$\mathbf{T}_a - \mathbf{I}$\\ $(0.001)$\end{tabular} &
		\begin{tabular}[c]{@{}l@{}}$\mathbf{0} \pm$\\ $5 \mathbf{1}_{3\times3}$\end{tabular} &
		\begin{tabular}[c]{@{}l@{}}-6.74 $\pm$ 1.69\\ 21.72 $\pm$ 5.08\\ -10.89 $\pm$ 3.09\\ -25.13 $\pm$ 4.36\\ -4.35 $\pm$ 2.26\\ 22.43 $\pm$ 6.61\\ 20.23 $\pm$ 3.77\\ -30.40 $\pm$ 4.16\\ 20.70 $\pm$ 5.04\end{tabular} &
		\begin{tabular}[c]{@{}l@{}}-7.76 $\pm$ 1.91\\ 25.82 $\pm$ 3.08\\ -9.58 $\pm$ 3.33\\ -27.79 $\pm$ 3.46\\ -3.92 $\pm$ 2.20\\ 29.00 $\pm$ 5.22\\ 18.90 $\pm$ 2.94\\ -35.93 $\pm$ 3.76\\ 23.66 $\pm$ 3.00\end{tabular} &
		\begin{tabular}[c]{@{}l@{}}-4.67\\ 29.59\\ -7.41\\ -29.32\\ -2.39\\ 34.55\\ 18.71\\ -34.10\\ 29.75\end{tabular} \\ \hline
		\begin{tabular}[c]{@{}l@{}}$\mathbf{p}_{C0B}$\\ $(cm)$\end{tabular} &
		$\mathbf{0} \pm 2\mathbf{1}$ &
		\begin{tabular}[c]{@{}l@{}}3.82 $\pm$ 0.91\\ -5.40 $\pm$ 0.67\\ -4.57 $\pm$ 1.31\end{tabular} &
		\begin{tabular}[c]{@{}l@{}}3.82 $\pm$ 0.32\\ -4.60 $\pm$ 0.55\\ -8.00 $\pm$ 0.46\end{tabular} &
		\begin{tabular}[c]{@{}l@{}}4.71\\ -4.76\\ -6.90\end{tabular} \\ \hline
		\begin{tabular}[c]{@{}l@{}}$f^0, c_x^0, c_{y}^0$\\ $(px)$\end{tabular} &
		\begin{tabular}[c]{@{}l@{}}$0 \pm 5$\\ $0 \pm 5$\\ $0 \pm 5$\end{tabular} &
		\begin{tabular}[c]{@{}l@{}}2.72 $\pm$ 0.87\\ -0.58 $\pm$ 1.08\\ 1.53 $\pm$ 0.59\end{tabular} &
		\begin{tabular}[c]{@{}l@{}}0.00 $\pm$ 0.00\\ 0.00 $\pm$ 0.00\\ 0.00 $\pm$ 0.00\end{tabular} &
		\begin{tabular}[c]{@{}l@{}}0.98\\ -1.07\\ 0.90\end{tabular} \\ \hline
		\begin{tabular}[c]{@{}l@{}}$k_1^0, k_2^0$\\ $k_3^0, k_4^0$\\ $(0.001)$\end{tabular} &
		\begin{tabular}[c]{@{}l@{}}$0 \pm 50$\\ $0 \pm 10$\\ $0 \pm 1$\\ $0 \pm 1$\end{tabular} &
		\begin{tabular}[c]{@{}l@{}}0.62 $\pm$ 1.89\\ -1.22 $\pm$ 1.01\\ -0.18 $\pm$ 0.13\\ -0.17 $\pm$ 0.06\end{tabular} &
		\begin{tabular}[c]{@{}l@{}}0.00 $\pm$ 0.00\\ 0.00 $\pm$ 0.00\\ 0.00 $\pm$ 0.00\\ 0.00 $\pm$ 0.00\end{tabular} &
		\begin{tabular}[c]{@{}l@{}}3.48\\ 0.72\\ -2.05\\ 0.20\end{tabular} \\ \hline
		$t_d^0 (ms)$ &
		$0 \pm 5$ &
		0.25 $\pm$ 0.08 &
		0.21 $\pm$ 0.09 &
		0.13 \\ \hline
		\begin{tabular}[c]{@{}l@{}}$\mathbf{p}_{BC1}$\\ $(cm)$\end{tabular} &
		$\mathbf{0} \pm 2\mathbf{1}$ &
		\begin{tabular}[c]{@{}l@{}}-4.48 $\pm$ 1.92\\ -4.01 $\pm$ 2.14\\ -5.24 $\pm$ 0.67\end{tabular} &
		\begin{tabular}[c]{@{}l@{}}-6.48 $\pm$ 0.32\\ -7.61 $\pm$ 0.42\\ -5.66 $\pm$ 0.30\end{tabular} &
		\begin{tabular}[c]{@{}l@{}}-5.40\\ -7.01\\ -4.95\end{tabular} \\ \hline
		$\delta\boldsymbol{\theta}_{BC1} ({}^{\circ})$ &
		\begin{tabular}[c]{@{}l@{}}$\mathbf{0} \pm$\\ $0.57 \mathbf{1}$\end{tabular} &
		\begin{tabular}[c]{@{}l@{}}-2.62 $\pm$ 0.17\\ 0.03 $\pm$ 0.07\\ -0.26 $\pm$ 0.48\end{tabular} &
		\begin{tabular}[c]{@{}l@{}}-3.18 $\pm$ 0.10\\ -0.03 $\pm$ 0.02\\ 0.74 $\pm$ 0.11\end{tabular} &
		\begin{tabular}[c]{@{}l@{}}-2.69\\ -0.05\\ 0.04\end{tabular} \\ \hline
		\begin{tabular}[c]{@{}l@{}}$f^1, c_{x}^1, c_{y}^1$\\ $(px)$\end{tabular} &
		\begin{tabular}[c]{@{}l@{}}$0 \pm 5$\\ $0 \pm 5$\end{tabular} &
		\begin{tabular}[c]{@{}l@{}}2.47 $\pm$ 1.39\\ -5.02 $\pm$ 1.89\\ -0.56 $\pm$ 0.86\end{tabular} &
		\begin{tabular}[c]{@{}l@{}}0.00 $\pm$ 0.00\\ 0.00 $\pm$ 0.00\\ 0.00 $\pm$ 0.00\end{tabular} &
		\begin{tabular}[c]{@{}l@{}}0.44\\ -3.40\\ -1.08\end{tabular} \\ \hline
		\begin{tabular}[c]{@{}l@{}}$k_1^1, k_2^1$\\ $k_3^1, k_4^1$\\ $(0.001)$\end{tabular} &
		\begin{tabular}[c]{@{}l@{}}$0 \pm 50$\\ $0 \pm 10$\\ $0 \pm 1$\\ $0 \pm 1$\end{tabular} &
		\begin{tabular}[c]{@{}l@{}}2.59 $\pm$ 3.29\\ -2.13 $\pm$ 2.31\\ -0.27 $\pm$ 0.15\\ -0.06 $\pm$ 0.16\end{tabular} &
		\begin{tabular}[c]{@{}l@{}}0.00 $\pm$ 0.00\\ 0.00 $\pm$ 0.00\\ 0.00 $\pm$ 0.00\\ 0.00 $\pm$ 0.00\end{tabular} &
		\begin{tabular}[c]{@{}l@{}}3.40\\ 1.77\\ -2.66\\ 0.33\end{tabular} \\ \hline
		$t_d^1 (ms)$ &
		$0 \pm 5$ &
		0.24 $\pm$ 0.17 &
		0.20 $\pm$ 0.11 &
		0.13 \\ \hline
	\end{tabular}
\end{table}

By comparing the estimated sensor parameters by KSF and reference
values in Table~\ref{tab:tumvi-room-calib} (third column versus last one),
we found that parameters $\{\mathbf{T}_g, \mathbf{T}_s\}$ could be estimated
very close to their reference values with small uncertainty. 
The small estimates for $\mathbf{T}_s\}$ entries mean that the g-sensitivity is negligible for the IMU of TUM VI.
Estimates for entries of $\{\mathbf{T}_a\}$ had more uncertainty, 
likely because they are more challenging to calibrate.
Some IMU bias estimates were not very close to reference values mainly because
biases are typically random walk processes and 
reference values are obtained from independent calibration data.

The camera extrinsic and intrinsic parameters, \eg, $\mathbf{p}_{C0B}$, 
$f^0, c_x^0, c_y^0$ were mostly estimated with the uncertainty range $[-\sigma, \sigma]$ covering their reference values.
The distortion parameters, \eg, $k_1^0, k_2^0, k_3^0, k_4^0$, often had large
uncertainty and even obvious deviations from reference values at high order
terms. This likely indicated these parameters were not well observable.

When running KSF on these raw sequences, locking the IMU intrinsic parameters
$\{\mathbf{T}_g, \mathbf{T}_s, \mathbf{T}_a\}$ caused estimated poses of most
sequences to drift more than 1000 m. By contrast, KSF calibrating all parameters
except frame readout time achieved RRE 0.136 $^{\circ}/m$ and RTE 0.637\% on the
raw room sequences.

If the camera intrinsic parameters were known well close
to their true values, it might be tempting to lock these parameters during
online calibration. With this setting, calibration results for raw room
sequences are given in Column 4, Table~\ref{tab:tumvi-room-calib}.
Compared to calibrating all parameters (Column 3, Table~\ref{tab:tumvi-room-calib}),
the estimates for $\{\mathbf{T}_g, \mathbf{T}_s, \mathbf{T}_a\}$ and camera
extrinsic parameters generally came closer to their reference values.
Also, in this case, KSF achieved RRE 0.101 $^{\circ}/m$ and RTE 0.345\%
slightly better than KSF calibrating all parameters.
This indicated that calibrating camera intrinsic parameters beforehand and
locking them at good estimates can help calibrating other parameters, \eg, IMU
intrinsic parameters.

\textbf{Temporal Parameter Estimation}
For calibrating frame readout time, we tested KSF with three 10-minute sequences
collected with rear color cameras and IMUs on three Android smartphones,
Honor V10, Lenovo Phab2 Pro, and Asus ZenFone \cite{huai_mobile_2019} which were moved randomly but smoothly in our workplace.
We split a sequence
into four segments, each about two minutes, then ran KSF three times on each
segment, estimating all sensor parameters as in the Torus case of Table~\ref{tab:sim_state_estimate}.
For all runs, time offset and frame readout time of the camera were initialized to 0 ms and 30 ms.
The state vectors estimated at the end of these 12 runs for a smartphone
were used to compute mean and standard deviation of time delay and frame
readout time of the camera.
For comparison, we also computed the frame readout time, \ie, rolling
shutter skew, with a LED panel which turns on columns of LEDs
sequentially. By examining the tilted line caused by the rolling
shutter effect on an image, the frame readout time can be easily computed \cite{LEDPanelV4User2020}.
For each smartphone, about 100 images captured at varying exposure
durations were used to compute the mean and standard deviation of the
frame readout time given in Table~\ref{tab:temporal-params}.
From the table, we see
that the time delay can be estimated with very small uncertainty.
The frame readout times by KSF, by the LED panel method, and from Android API are reasonably close.
Interestingly, estimates by KSF are consistently slightly smaller than
those obtained with the LED panel.
\begin{table}[]
	\centering
	\caption{Temporal parameters estimated by KSF for the rear color camera
		of smartphones including Honor V10, Lenovo Phab2 Pro, and Asus ZenFone,
		and frame readout time returned by the Android camera API and estimated
		by manual drawing on images of a LED panel.}
	\label{tab:temporal-params}
	\begin{tabular}{l|l|lll}
		\hline
		& \begin{tabular}[c]{@{}l@{}}Time offset\\ (ms)\end{tabular} & \multicolumn{3}{l}{Frame readout time (ms)} \\ \hline
		Phone     & KSF                 & KSF              & API   & LED panel        \\ \hline
		V10       & 12.946 $\pm$ 0.014  & 18.94 $\pm$ 0.15 & 20.73 & 20.68 $\pm$ 0.13 \\ \hline
		Phab2 Pro & -26.332 $\pm$ 0.038 & 29.34 $\pm$ 0.26 & 30.25 & 31.39 $\pm$ 0.30 \\ \hline
		ZenFone   & -3.743 $\pm$ 0.066  & 6.45 $\pm$ 0.12  & 5.47  & 7.41 $\pm$ 0.31  \\ \hline
	\end{tabular}
\end{table}

\subsection{Pose Estimation}
\label{subsec:pose-estimation}
With two public benchmarks for VIO, EuRoC~\cite{burri2016euroc}
captured by a MAV and TUM VI~\cite{schubert2018tum}
captured by a handheld device, we evaluated the performance of KSF in
regard to pose accuracy and compared to several recent open sourced VIO methods,
OKVIS~\cite{leutenegger_keyframe_2015}, OpenVINS~\cite{geneva_openvins_2019} 
and a stereo VIO method \cite{usenko_visual_2020}. The evaluation results of KSF versus other methods on
the UZH-FPV dataset~\cite{delmerico2019we} is available at its
website\footnote{https://fpv.ifi.uzh.ch/uzh/2019-2020-uzh-fpv-temporary-leader-board/osu-ethz-2/}.

These methods were tested in two sensor setups, the left camera and the IMU, and
the stereo cameras and the IMU, except for the stereo VIO method \cite{usenko_visual_2020} which only supports the latter.
To simplify tuning parameters, each method with a specific sensor setup
processed all data sequences of a benchmark with the same set of parameters
unless otherwise specified. Among parameters given to these methods,
we observed that the
IMU noise parameters often required manual tuning. If need be, the proper
values for these parameters were found by a simple search of a coefficient on
white noise $c_{n}$ and a coefficient on bias white noise $c_{bn}$ starting from
baseline values of $\{\sigma_g, \sigma_a, \sigma_{bg}, \sigma_{ba}\}$. 
The coefficient pair $(c_{n}, c_{bn})$ scales the noise parameters as 
\begin{equation}
\label{eq:scale-imu-noise}
\begin{split}
\sigma_g \leftarrow c_{n} \sigma_g &\quad 
\sigma_a \leftarrow c_{n} \sigma_a \\
\sigma_{bg} \leftarrow c_{bn} \sigma_{bg} &\quad
\sigma_{ba} \leftarrow c_{bn} \sigma_{ba}.
\end{split}
\end{equation}

For KSF, we used the IMU-centric body frame
(Table~\ref{tab:imu-camera-extrinsic-models}),
5 old keyframes and 5 temporal frames, \ie, $N_{kf} = 5$ and $N_{tf} = 5$, in
the sliding window, and locked camera intrinsic parameters and frame readout
times while estimating camera extrinsic parameters and time offsets. These
settings were sensible and mostly supported by the sensitivity analysis in
Section \ref{subsec:sensitivity}.
For OKVIS, OpenVINS, and the stereo VIO method \cite{usenko_visual_2020}, we configured them with as many
default parameters provided by their authors as possible.
Specifically, OpenVINS was configured to run synchronously with landmark variables in
the state vector. When data from two cameras were available, the stereo mode was
chosen over the binocular mode.
Method-specific parameters dependent on data sequences are detailed in the following test results.

All methods were executed synchronously without
enforcing real-time processing. This was specially relevant for OKVIS which often
ran nearly real-time on our test computer.
To reduce the random behavior, every method ran three times on every sequence.

\textbf{Benchmarking on EuRoC}
The EuRoC benchmark~\cite{burri2016euroc} was captured indoors by a MAV mounted with a sensor platform of
stereo cameras of global shutters and a 10-DOF IMU worth $\sim$\$500.
Ground truth of centimeter level position accuracy by a motion capture system
was provided covering all sequences.
For better comparison with other reported results on EuRoC, we used the ground
truth corrected by \cite{geneva_openvins_2019} in orientation for sequence
V1\_01 and otherwise the ground truth provided by EuRoC.
Additionally, sequence V2\_03 was excluded because its missing images caused
some algorithm to break.

To save tuning efforts, the same IMU noise parameters were used for OKVIS with
stereo cameras, KSF with a monocular camera, and KSF with stereo cameras.
However, for OKVIS with a monocular camera, we had to inflate IMU noise
parameters by the coefficient pair (4, 4) relative to the stereo OKVIS setting
due to a grid search \eqref{eq:scale-imu-noise} for expected pose accuracy.
For OpenVINS, stationary segments at the start of several
sequences were skipped to ensure good initialization.

For these methods with a monocular or stereo setup, RMS of ATE on each sequence
and overall relative errors are tabulated in
Table~\ref{tab:euroc-comparison}. 
OpenVINS had good results in the monocular setup.
The stereo VIO method \cite{usenko_visual_2020} achieved best results with stereo cameras. 
Overall, KSF achieved pose accuracy comparable to recent VIO methods on EuRoC
sequences.

\begin{table*}[]
	\centering
	\caption{Mean RMS of ATE in meters and overall relative errors of VIO methods on EuRoC dataset over three runs. Best scores in either monocular or stereo mode are in $\mathbf{bold}$. RRE: Relative Rotation Error, RTE: Relative Translation Error.}
	\label{tab:euroc-comparison}
	\begin{tabular}{l|l|llllllllllll}
		\hline
		\multicolumn{2}{l|}{Method} &
		MH\_01 &
		MH\_02 &
		MH\_03 &
		MH\_04 &
		MH\_05 &
		V1\_01 &
		V1\_02 &
		V1\_03 &
		V2\_01 &
		V2\_02 &
		\begin{tabular}[c]{@{}l@{}}RRE\\ ($^{\circ}/m$)\end{tabular} &
		\begin{tabular}[c]{@{}l@{}}RTE\\ (\%)\end{tabular} \\ \hline
		\multirow{3}{*}{mono} &
		OKVIS &
		0.23 &
		0.22 &
		\textbf{0.15} &
		0.37 &
		\textbf{0.34} &
		0.09 &
		0.08 &
		0.15 &
		\textbf{0.09} &
		0.11 &
		\textbf{0.024} &
		0.66 \\ \cline{2-14} 
		&
		OpenVINS &
		0.15 &
		\textbf{0.09} &
		\textbf{0.15} &
		\textbf{0.16} &
		0.48 &
		\textbf{0.06} &
		\textbf{0.06} &
		\textbf{0.07} &
		0.11 &
		\textbf{0.06} &
		0.028 &
		\textbf{0.50} \\ \cline{2-14} 
		&
		KSF &
		\textbf{0.13} &
		0.29 &
		0.16 &
		0.29 &
		0.42 &
		0.09 &
		0.11 &
		0.19 &
		0.12 &
		0.15 &
		0.027 &
		0.69 \\ \hline
		\multirow{4}{*}{stereo} &
		OKVIS &
		0.17 &
		0.16 &
		0.15 &
		0.20 &
		0.29 &
		0.03 &
		0.07 &
		0.13 &
		0.06 &
		0.09 &
		0.025 &
		0.52 \\ \cline{2-14} 
		&
		OpenVINS &
		0.10 &
		0.10 &
		0.20 &
		0.29 &
		0.21 &
		0.06 &
		0.07 &
		\textbf{0.05} &
		0.10 &
		\textbf{0.05} &
		0.030 &
		0.48 \\ \cline{2-14} 
		&
		Usenko \etal \cite{usenko_visual_2020} &
		0.09 &
		\textbf{0.06} &
		\textbf{0.09} &
		\textbf{0.11} &
		\textbf{0.11} &
		\textbf{0.03} &
		0.06 &
		0.08 &
		0.04 &
		0.06 &
		\textbf{0.022} &
		\textbf{0.28} \\ \cline{2-14} 
		&
		KSF &
		\textbf{0.05} &
		0.08 &
		0.10 &
		0.25 &
		0.27 &
		0.04 &
		\textbf{0.05} &
		0.09 &
		\textbf{0.04} &
		0.06 &
		0.027 &
		0.37 \\ \hline
	\end{tabular}
\end{table*}

\textbf{Benchmarking on TUM VI}
The TUM VI benchmark~\cite{schubert2018tum} were captured indoors and outdoors by a handheld device
including stereo cameras of global shutters and a MEMS IMU worth $\sim$\$4.
For each of eight outdoor sequences, the traveled distance was greater
than 900 m, the maximum being 2656 m.
Ground truth of centimeter level position accuracy by a
motion capture system was provided for all sequences at the start and at the
end, and ground truth covering the whole trajectory were available for
sequences collected in the motion capture room.

For OKVIS in both monocular and stereo modes, the configuration file provided by
TUM VI was used.
For OpenVINS, the acceleration variance threshold, \texttt{init\_imu\_thresh},
was adjusted individually for each sequence so that the algorithm could start
without apparent drifting.
Internally, the parameter is used to detect if the device is moving and thus the
algorithm is ready to be initialized.
For \cite{usenko_visual_2020}, parameters provided by the author were used.

For monocular or stereo KSF, the IMU noise parameters were tuned by the
aforementioned simple search starting from a baseline, \ie,
the noise parameters for OKVIS. For instance, with monocular KSF, we searched the
list of coefficient pairs with a roughly step size 2 on the logarithmic scale,
$\{(0.05, 0.1), (0.1, 0.1), (0.1, 0.25), (0.2, 0.25)\}$, which were selected
based on our observation that suitable IMU noise parameters for KSF tend to be smaller than those for OKVIS.
In the end, we found that monocular KSF worked well with coefficient pair 
(0.2, 0.25) and stereo KSF with (0.1, 0.1).

All methods processed all sequences without significant drift 
\textgreater 1000 m at the end.
The mean of RMS ATE over three runs of each sequence and 
overall relative errors are tabulated in Table~\ref{tab:tumvi-comparison}.
From Table~\ref{tab:tumvi-comparison}, we see that all methods
performed well with indoor sequences, but had difficulties with outdoor sequences.
Interestingly, all methods tackled well scenarios with few
features, \eg, slides1 - 3 which have segments captured while sliding through a tube.
From RMS ATE values, there is no clear winner, 
as the best scores are distributed among these methods.
From relative errors, KSF in both monocular and stereo modes
achieved overall best RTE.

\begin{table}[]
	\centering
	\caption{Mean RMS of ATE in meters and overall relative errors of VIO methods on TUM-VI benchmark over three runs. Best scores in either monocular or stereo mode are in $\mathbf{bold}$. RRE: Relative Rotation Error, RTE: Relative Translation Error, cor: corridor, mag:magistrale, out: outdoors.}
	\label{tab:tumvi-comparison}
	\begin{tabular}{l|lll|llll}
		\hline
		\multirow{2}{*}{Session} & \multicolumn{3}{c|}{Mono}                        & \multicolumn{4}{c}{Stereo}                                      \\ \cline{2-8} 
		&
		\begin{tabular}[c]{@{}l@{}}OK-\\ VIS\end{tabular} &
		\begin{tabular}[c]{@{}l@{}}Open\\ VINS\end{tabular} &
		KSF &
		\begin{tabular}[c]{@{}l@{}}OK-\\ VIS\end{tabular} &
		\begin{tabular}[c]{@{}l@{}}Open\\ VINS\end{tabular} &
		\begin{tabular}[c]{@{}l@{}}\cite{usenko_visual_2020}\end{tabular} &
		KSF \\ \hline
		cor1                     & 1.08           & \textbf{0.49}  & 0.74           & \textbf{0.12}  & 0.47           & 0.47          & 0.79           \\ \hline
		cor2                     & 0.90           & 0.83           & \textbf{0.60}  & 0.59           & 1.07           & 0.55          & \textbf{0.46}  \\ \hline
		cor3                     & 0.78           & \textbf{0.28}  & 0.58           & \textbf{0.42}  & 0.47           & 0.46          & 0.74           \\ \hline
		cor4                     & \textbf{0.30}  & 0.80           & 0.33           & 0.31           & 0.41           & 0.25          & \textbf{0.12}  \\ \hline
		cor5                     & 0.46           & \textbf{0.23}  & 0.37           & 0.44           & \textbf{0.32}  & 0.40          & 0.42           \\ \hline
		mag1                     & 5.73           & \textbf{1.32}  & 2.75           & 4.69           & 3.40           & \textbf{1.21} & 1.73           \\ \hline
		mag2                     & 10.20          & \textbf{3.12}  & 12.78          & 1.90           & 2.40           & \textbf{1.11} & 10.18          \\ \hline
		mag3                     & \textbf{0.98}  & 4.36           & 1.22           & 0.98           & 2.32           & \textbf{0.74} & 1.09           \\ \hline
		mag4                     & 5.64           & \textbf{3.47}  & 4.38           & \textbf{0.65}  & 3.50           & 1.59          & 8.10           \\ \hline
		mag5                     & \textbf{1.30}  & 2.38           & 2.07           & 1.33           & 2.98           & \textbf{0.61} & 1.81           \\ \hline
		mag6                     & 8.04           & \textbf{2.93}  & 11.93          & 4.46           & \textbf{1.86}  & 3.43          & 2.10           \\ \hline
		out1                     & 104.3          & 108.8          & \textbf{69.63} & 93.14          & 118.6          & 255.1         & \textbf{80.68} \\ \hline
		out2                     & 109.5          & \textbf{37.39} & 51.65          & 49.91          & \textbf{20.90} & 64.63         & 29.41          \\ \hline
		out3                     & \textbf{17.40} & 40.44          & 45.15          & \textbf{22.14} & 54.56          & 38.26         & 25.56          \\ \hline
		out4                     & 23.10          & 43.73          & \textbf{7.41}  & 22.43          & 25.32          & 17.54         & \textbf{8.89}  \\ \hline
		out5                     & 17.81          & 78.06          & \textbf{9.45}  & \textbf{7.67}  & 12.81          & 7.89          & 13.09          \\ \hline
		out6                     & 138.2          & \textbf{46.18} & 80.83          & 85.51          & 87.88          & 65.53         & \textbf{22.03} \\ \hline
		out7                     & \textbf{20.41} & 22.95          & 31.16          & 4.61           & 14.37          & \textbf{4.16} & 30.86          \\ \hline
		out8                     & \textbf{3.21}  & 31.51          & 7.32           & \textbf{5.64}  & 12.03          & 13.54         & 11.05          \\ \hline
		room1                    & 0.09           & \textbf{0.06}  & 0.08           & 0.06           & 0.06           & 0.09          & \textbf{0.05}  \\ \hline
		room2                    & 0.13           & \textbf{0.06}  & 0.20           & 0.10           & \textbf{0.07}  & \textbf{0.07} & \textbf{0.07}  \\ \hline
		room3                    & \textbf{0.07}  & 0.10           & 0.09           & \textbf{0.06}  & 0.08           & 0.13          & 0.07           \\ \hline
		room4                    & \textbf{0.04}  & \textbf{0.04}  & 0.23           & \textbf{0.04}  & \textbf{0.04}  & 0.06          & \textbf{0.04}  \\ \hline
		room5                    & \textbf{0.08}  & \textbf{0.08}  & 0.11           & \textbf{0.06}  & 0.08           & 0.14          & 0.10           \\ \hline
		room6                    & 0.08           & 0.09           & \textbf{0.05}  & 0.04           & 0.04           & \textbf{0.02} & 0.03           \\ \hline
		slides1                  & 2.12           & \textbf{0.28}  & 0.43           & \textbf{0.33}  & 0.84           & \textbf{0.33} & 1.26           \\ \hline
		slides2                  & 4.37           & \textbf{1.97}  & 3.47           & 1.30           & \textbf{1.08}  & 0.35          & 1.34           \\ \hline
		slides3                  & 2.80           & \textbf{0.91}  & 3.03           & \textbf{0.87}  & 1.66           & 0.90          & 2.86           \\ \hline
		\begin{tabular}[c]{@{}l@{}}RRE\\ ($^{\circ}/m$)\end{tabular} &
		0.156 &
		0.120 &
		\textbf{0.118} &
		\textbf{0.087} &
		0.105 &
		0.118 &
		0.135 \\ \hline
		\begin{tabular}[c]{@{}l@{}}RTE\\ (\%)\end{tabular} &
		58.47 &
		57.87 &
		\textbf{42.11} &
		45.70 &
		48.52 &
		63.64 &
		\textbf{31.93} \\ \hline
	\end{tabular}
\end{table}

\subsection{Computation Timing}
\label{subsec:timing}
We measured the time consumption of major components of KSF on a consumer
laptop, Dell Insipiron 7590. 
The laptop had 16 GB RAM, a Intel Core i7-9750H processor with 6 cores of
base frequency 2.60 GHz.
The program of KSF ran on Ubuntu 18.04 without GPU acceleration.
In implementation, the KSF program is divided into three stages of a processing
pipeline, feature extraction, feature matching, and estimation including filter
update and marginalization. Each frame goes through these three ordered stages
to complete processing. Ideally these stages run in parallel while working on
different frames such that the throughput is only restrained by the most time-consuming stage. 
However, the feature matching stage depends on results of the estimation stage (Fig.~\ref{fig:flowchart}).
Thus, the average processing time for a frame is the
total time for feature matching and estimation.

To time these stages, monocular or stereo KSF was used to synchronously
process the whole EuRoC dataset once with timing service on.
All sensor parameters were unlocked, \ie, IMU 
parameters \eqref{eq:IMU-parameters}, camera extrinsic, intrinsic and temporal
parameters, $t_d^k$ and $t_r^k$, for $N$ cameras.
In the sliding window, we kept $N_{kf} = 7$ keyframes and 
$N_{tf} = 5$ recent frames.
From an image of a frame, up to 400 keypoints were extracted.

\begin{table}[]
	\centering
	\caption{Time cost of major functions of KSF in processing EuRoC sequences In milliseconds. Values in $\textbf{bold}$ identity routines that determine average time to process a frame.}
	\label{tab:timing}
	\begin{tabular}{l|ll|ll}
		\hline
		\multicolumn{1}{c|}{\multirow{2}{*}{Function}} & \multicolumn{2}{c|}{Mono} & \multicolumn{2}{c}{Stereo} \\ \cline{2-5} 
		\multicolumn{1}{c|}{} & mean         & $\sigma$ & mean          & $\sigma$ \\ \hline
		Feature extraction 0  & 3.0          & 0.3      & 5.1           & 1.5      \\
		Feature extraction 1  & 0            & 0        & 5.1           & 1.6      \\ \hline
		Match to keyframes    & \textbf{1.3} & 0.2      & \textbf{2.9}  & 0.7      \\
		Match to last frame   & \textbf{0.7} & 0.1      & \textbf{1.8}  & 0.5      \\
		Match left to right   & 0            & 0        & \textbf{0.5}  & 0.1      \\ \hline
		Filter update         & \textbf{4.5} & 1.4      & \textbf{13.4} & 3.6      \\
		Marginalization       & \textbf{1.5} & 0.8      & \textbf{5.1}  & 1.9      \\ \hline
		Time per frame        & \textbf{8.0} & N/A      & \textbf{23.7} & N/A      \\ \hline
	\end{tabular}
\end{table}

Means and standard deviations of computation costs for major components are
given in Table~\ref{tab:timing}.
From the table, it is easy to see that based on the pipelining
discussion, the average time to process a frame was 8 ms (125 Hz) in the monocular
mode, and 23.7 ms (42.2 Hz) in the stereo mode.
These numbers show that KSF can process the EuRoC dataset in real time
on a consumer laptop.
It is worth noting that feature matching functions of stereo KSF, \eg, match to
last frame, took at least twice as long as the counterparts of monocular KSF
because of data from two cameras. 
And the filter update and marginalization steps of stereo KSF were about three times
as expensive as those of monocular KSF. This was probably caused by
increased number of feature tracks due to stereo matching and parameters
added to the state vector for the extra camera.
To put this in perspective, KSF runs at least three times as fast as OKVIS,
faster than OpenVINS because landmarks are not included in the state vector,
slower than the stereo VIO method \cite{usenko_visual_2020} which utilizes parallel implementation.

\section{Discussion}
\label{sec:discussion}
The KSF method is versatile enough to calibrate a variety of sensor parameters,
including IMU scaling and misalignment errors and camera intrinsic parameters.
As expected, calibrating these parameters requires a rich variety of motion that renders them observable.
Our simulation in Section \ref{tab:sim_state_estimate} showed that
estimating all sorts of IMU parameters caused slow convergence of biases when the
device went through monotonic motion. Additionally, tests on real-world data
indicated that estimating camera intrinsic parameters online led to slightly
worse pose estimation (Table~\ref{tab:sensitivity-calibration}, and
Section~\ref{subsec:sensor-calibration}) when we had good estimates for these
parameters. In such cases, these less observable or well calibrated parameters
are better locked or alternative sensor models
(Table~\ref{tab:imu-camera-extrinsic-models}) can be used.

KSF handles standstill periods at the start or in the middle of execution without
distinguishing motion types or using zero velocity update. 
The main reason is the use of keyframes which typically do not change
when the platform is stationary.
A minor reason is that KSF uses feature tracks of low disparities to update the
estimator as done in OKVIS~\cite{leutenegger_keyframe_2015}. Such feature tracks
appear more frequently when the camera platform is stationary or goes through pure
rotation. The corresponding landmarks are initialized with a large depth, \eg,
1000 m, and used in the filter update step. Thus, KSF can initialize even
when the platform is stationary. Obviously doing so trades accuracy
for robustness and fast initialization.

Empirically, KSF tackled well scenes with few features in TUM VI
\cite{schubert2018tum}, glass elevators in ADVIO~\cite{advio_2018}.
But it had a large jump in position, $\sim$70 m, for an ADVIO sequence captured
inside a steel elevator, and it often under estimated the vertical motion for
ADVIO sequences captured on escalators. These cases are challenging because the
apparent stationary visual features are nearly or literally 100\% outliers.
Unsurprisingly, many proprietary algorithms also broken down \cite{advio_2018}.
In the end, we recommend turning off VIO methods in these special scenarios.

Furthermore, we observed that when the $N$-camera-IMU system was mounted on a ground
vehicle which traversed rough terrain,
the accelerometer readings were so disturbed by vibration that a VIO estimator
did not give meaningful results. In such cases, measures to reduce vibration,
\eg, anti-vibration mounts \cite{titterton2004strapdown}, are often necessary.

\section{Conclusion And Future Work}
\label{sec:conclusion}
This paper presents a versatile Keyframe-based Structureless Filter (KSF) for
visual inertial odometry with one or more cameras.
The method associates features relative to keyframes in the frontend and
manages navigation state variables in the filter's sliding window according to
whether they are associated with keyframes, thus the method is able to deal with
stationary scenarios.
This framework accommodates a couple of IMU models,
several camera extrinsic models, and a number of
pinhole projection and lens distortion models.
The generic design of the filter allows calibrating or locking scaling,
misalignment, and relative orientation errors of the IMU, and intrinsic, 
extrinsic, and temporal parameters of individual cameras. 

The filter is able to estimate the sensor parameters accurately when the motion
is diverse enough for the sensor setup. This was confirmed with simulation and
by estimating the full spectrum of sensor parameters for raw sequences of 
TUM VI benchmark~\cite{schubert2018tum}.
From these tests, we observed that it might benefit state estimation to lock some
systematic errors of the IMU when the motion is monotonic or
to lock the camera intrinsic parameters when the camera is well calibrated.

Also, KSF tracks device motion precisely with consistent covariance estimates as
shown by simulation. On EuRoC and TUM VI benchmarks, KSF with online calibration
of camera extrinsic parameters and time offsets, achieved competitive pose
accuracy relative to recent VIO methods,
OKVIS~\cite{leutenegger_keyframe_2015},
OpenVINS~\cite{geneva_openvins_2019}, and
a stereo VIO method~\cite{usenko_visual_2020}. In terms of speed, KSF processed EuRoC
stereo image streams at 42 Hz while performing full calibration.

In the near future, we will study alternative error state for filtering, 
\eg, right invariant errors \cite{zhang_invariance_2018}, which guarantees
estimator consistency without manipulating Jacobians.

Another interesting research topic is how to easily adapt a VIO method to data
collected by different sensors. In obtaining the accuracy report on public
benchmarks for the compared VIO methods, much effort was dedicated to adjusting
IMU noise parameters because of different sensors and IMU integration
implementations. As a result, we look forward to an approach that alleviates 
the "curse of manual tuning" \cite{cadena2016past}.

\ifCLASSOPTIONcaptionsoff
  \newpage
\fi

\bibliographystyle{IEEEtran}
\bibliography{IEEEabrv,ms}

\end{document}